\pdfoutput=1
\documentclass[11pt]{article}

\usepackage[preprint]{acl}
\definecolor{Gray}{gray}{0.5}
\usepackage{times}
\usepackage{latexsym}
\usepackage{array}
\usepackage{microtype}
\usepackage{graphicx}
\usepackage{diagbox}
\usepackage{bbold}
\usepackage{multirow}
\usepackage{subcaption}
\usepackage{color}
\usepackage{booktabs} % for professional tables
\usepackage[T1]{fontenc}
\usepackage[utf8]{inputenc}

\usepackage{microtype}

\usepackage{inconsolata}

\usepackage{graphicx}
\usepackage{hyperref}       % hyperlinks
\usepackage{url}         
\usepackage{amsmath}
\usepackage{amssymb}
\usepackage{mathtools}
\usepackage{amsthm}
\usepackage{dsfont}
\usepackage{cleveref} 
\usepackage{booktabs}       % professional-quality tables
\usepackage{amsfonts}       % blackboard math symbols
\usepackage{nicefrac}       % compact symbols for 1/2, etc.
\usepackage{microtype}      % microtypography
\usepackage{xcolor}         % colors
\usepackage{subcaption}

\theoremstyle{plain}
\newtheorem{theorem}{Theorem}[section]

\theoremstyle{definition}
\newtheorem{definition}[theorem]{Definition}

\theoremstyle{remark}

\usepackage{graphicx}
\usepackage{color}
\usepackage{mathrsfs} % mathscr
\usepackage{wrapfig}
\makeatletter
\@namedef{ver@everyshi.sty}{}
\makeatother
\usepackage{tikz}
    \definecolor{mycolor1}{HTML}{298562}%{009900}
    \definecolor{mycolor2}{HTML}{000000}
    \definecolor{mycolor3}{HTML}{A21329}%222cec}%{800000}
    \definecolor{mycolor4}{HTML}{222cec}%{000080}
  \definecolor{mycolor5}{HTML}{FFFF00}%
    \definecolor{mycolor6}{HTML}{990099}%
        \usetikzlibrary{arrows}
        \usetikzlibrary{decorations.markings}
        \tikzset{
                latex-arrow/.style={
                        decoration={markings,mark=at position 1 with {\arrow[scale=1.5,#1]{latex}}},
                        postaction={decorate},
                        shorten >=0.4pt
                },
                latex-arrow/.default=black
        }
        \tikzset{
                latex-arrow-red/.style={
                        decoration={markings,mark=at position 1 with {\arrow[scale=1.5,#1]{latex}}},
                        postaction={decorate},
                        shorten >=0.4pt
                },
                latex-arrow-red/.default=mycolor1
        }
        \tikzstyle{dotnode}=[circle,fill=black,inner sep=0ex,minimum size=1.5ex]
        \tikzstyle{mynode}=[circle, draw=black, fill=white, inner sep=0pt, minimum size=2ex]
        \tikzstyle{edge}=[draw=black,latex-arrow]
        \tikzstyle{red-edge}=[draw=mycolor1,latex-arrow-red]
\usepackage{microtype}
\title{Balancing Diversity and Risk in LLM Sampling: 
How to Select Your Method and Parameter for Open-Ended Text Generation}

\author{Yuxuan Zhou$^{1,2}$, Margret Keuper$^{1}$, Mario Fritz$^{2}$\\
  $^{1}$University of Mannheim, Germany \\
  $^{2}$CISPA Helmholtz Center for Information Security, Germany \\
  \texttt{yuxuan.zhou@cispa.de}, \texttt{keuper@uni-mannheim.de}, \texttt{fritz@cispa.de} 
  \\}

\begin{document}

\maketitle
\begin{abstract}
Sampling-based decoding strategies have been widely adopted for Large Language Models (LLMs) in numerous applications, targeting a balance between diversity and quality via temperature tuning and tail truncation. Considering the strong dependency 
% resolved
% \mario{"high dynamic range" seems to be the wrong wording here. i gueess you want to say "strong dependence on context" or "high variance in probability"? or better "strong variance in length of the tail as well as probability ranges"?} 
of the candidate next tokens on different prefixes, recent studies propose to adaptively truncate the tail of LLMs' predicted distribution. Although improved results have been reported with these methods on open-ended text generation tasks, the results are highly dependent on the curated parameters and the limited exemplar text. In this paper, we propose a systematic way to estimate the capacity 
% no good solution yet
% \mario{meaning of "intrinsic capacity" is not clear to the reader at this point} 
of a truncation sampling method by considering the trade-off between diversity and risk at each decoding step,
based on our collected prefix tree which preserves the context of a full sentence. 
% resolved
% \mario{last sentnec is not clear - but key. too long. tring to say too much at a time.} 
Our work offers a comprehensive comparison of existing truncation sampling methods and serves as a practical user guideline for their parameter selection.
Our code is available at \href{https://github.com/ZhouYuxuanYX/Benchmarking-and-Guiding-Adaptive-Sampling-Decoding-for-LLMs}{github repository}.
\end{abstract}

\section{Introduction}
\label{sec:intro}

Large Language Models (LLMs) \cite{achiam2023gpt, touvron2023llama, jiang2023mistral, team2023gemini} have demonstrated exceptional performance across a variety of applications, and the reliability of decoding strategies has become a critical concern. Previous works have revealed that likelihood-maximization such as beam search \cite{fan2018hierarchical, holtzman2019curious, welleck2019neural, meister2022probability} produces degenerate text which contains repetitive loops and incoherent context, particularly in open-ended tasks. Therefore, sampling-based decoding strategies, e.g., Top-p \cite{holtzman2019curious} and Top-k sampling \cite{radford2018improving, fan2018hierarchical}, have been widely adopted. The balance between diversity and quality of the generated text could be adjusted by tuning the temperature and truncation position to some extend, but requires non-trivial trial and error.

Recent studies \cite{basu2020mirostat, zhu2024improving, hewitt2022truncation, meister2023locally} proposed adaptive tail truncation mechanisms based on different criteria or assumptions, which maintain an allowed set of tokens with a flexible size according to the given prefix. To validate the effectiveness of a sampling method, they are often compared through extrinsic evaluation based on open-ended text generation applications. For example, story generation \cite{fan2018hierarchical} and document continuation \cite{merity2016pointer}. Various metrics \cite{welleck2019neural, meister2023locally, pillutla2021mauve, gao2021simcse} have been adopted to consider different aspects of the generated text.

We reveal two underlying issues in the current evaluation, which hinder the assessment of a method's significance in real-world applications:
\begin{itemize}
    \item \textbf{The improvement of one method over another might be simply due to a better tuned parameter for the targeted task}:
    the performance of sampling methods is sensitive to their parameters, and parameter sweep is often operated on a extremely sparse grid due to the high computation cost. This is especially problematic considering the non-linear dependency between performance and parameters.
    \item \textbf{Users are agnostic to the optimal parameters in real-world applications}: Practically speaking, users often pick parameters based on their own need for the compromise between diversity and quality, after a few tryouts. There exists no universal optimal paramters in different scenarios and users are agnostic to the optimal parameters for their own tasks.  
    
\end{itemize}

The above issues exactly indicate the need for an evaluation that allows for estimating the theoretical capacity of a truncation sampling method (how well it adapts to the variation in data supports given different prefixes), independent of hyperparamter tuning. Moreover, the second issue additionally highlights the need to identify the sweet spots of existing sampling methods, which could serve as a user guideline for practitioners. 

\begin{figure}[t]
    \centering
    \includegraphics[width=0.85\linewidth]{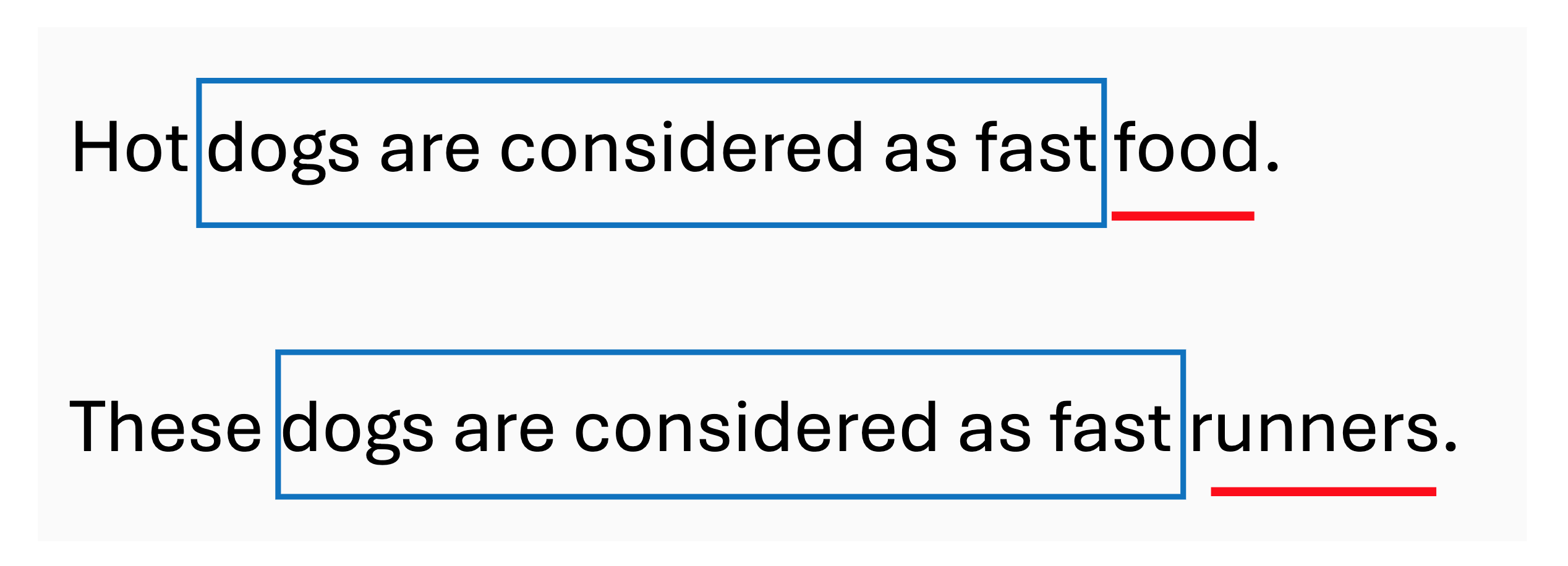}
    \caption{N-gram models tend to overestimate the data support size given a prefix (marked by a red line) due to limited window size (marked with a blue window).}
    \label{fig:example}
\end{figure}

In light of the above analysis, we propose a systematic way to assess the inherent adaptability of a sampling method. First, we rearrange Wikipedia-English \footnote{\url{https://dumps.wikimedia.org/}} data into a word-level prefix tree, known as a Trie \cite{fredkin1960trie,ghasemi2019granularity}. 
% resolved
% \mario{roughly from here until "given the CP-Trie" is too much information for an intro for my taste. explaining the whole algorithm in detail interrupts the folow in the intro. it also does not really tease the method section. I'm advising to keep it on a more conceptual level and reserver most of the technical details for later in the paper. the mission of the intro to make sure the reader understands the conceptional contribution -- everything else is baggage and rather a distraction from the main message.} 
It is noteworthy that a n-gram Trie \cite{jurafsky2000speech} tends to overestimate the data support size given a prefix \cite{bengio2000neural}, as shown in \Cref{fig:example}. In a similar spirit to \cite{ding2024fewer}, we construct the prefix tree with only sentence-starting n-grams to preserve full sentence context, called Context-Preserving Trie (CP-Trie). 

% Since the child nodes of a prefix define the data support, the recall of a sampling method w.r.t. data support can be computed. 

\begin{figure}
    \centering
    \includegraphics[trim={0cm 0cm 0cm 1cm}, width=0.45\textwidth]{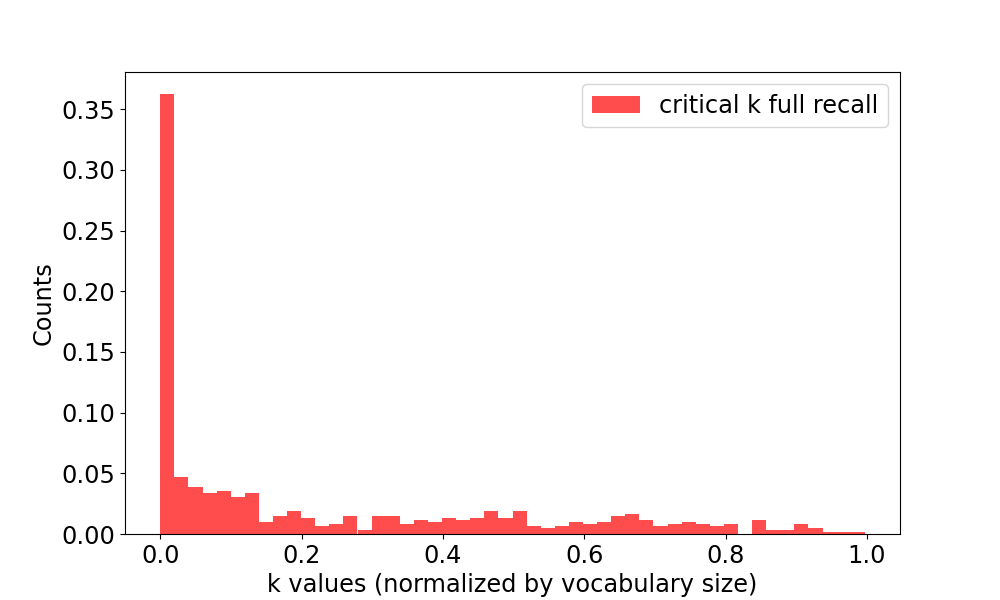}
   \caption{Histogram of the estimated optimal truncation values for gpt2-xl, which achieve exactly full recall of data support given different prefixes. 
    % The truncation p value of Top-p sampling is normalized to $[0, 1]$ for a direct comparison with Top-k sampling.
    }
    \label{fig:critical}
\end{figure}

Given the CP-Trie, we are able to estimate the theoretical capacity of a sampling method, by examining the amount of tokens within and out of the data support with varying truncation parameter values. 
% By restricting the evaluation to sentence-level context,  our data support serves as a reasonable lower bound despite the limited size of Wikipedia dataset, please see \Cref{sec:metric} for more details. 
As shown in \Cref{fig:critical}, the truncation positions, which exactly cover the full data supports,  vary drastically given different prefixes and Top-k sampling could be regarded as a baseline method with zero adaptability. Therefore, an adaptive truncation method is supposed to better follow such a variation, so that improved diversity can be achieved without harming the quality.

In summary, our contributions are as follows:
\begin{itemize}
 % \item We propose CP-Trie, which only consists of the sentence-starting n-grams to preserve the sentence-level context.
    \item We establish an intrinsic evaluation benchmark based on the collected CP-Trie, which allows for estimating the theoretical capacity of different sampling methods via thoroughly designed diversity and stability metrics.
    
    \item We conduct a comprehensive comparison of existing sampling approaches, which serves as a guideline for choosing a method and its  parameter in real-world applications. 

    \item We reveal that sampling-based decoding methods are underestimated in the existing study \citep{shi2024thorough} due to the difficulty in parameter selection, highlighting the merit of our evaluation protocol. 
    % resolved
    % \mario{I'm not sure if it wise to put the finger on this particular paper so badly. it almost sounds that you are saying that this paper is flawed. you could say "issue" instead of "pitfall". maybe the better alternative would be to phrase it in positive way - what we do right. and say that incomparison to other methods (cite) we have the following advantage.}
\end{itemize}

\section{Related Work}
In this section, we summarize recent sampling decoding strategies, along with common benchmarks and metrics for open-ended text generation.

\subsection{Sampling-based Decoding Methods}
Vanilla sampling suffers from the risk of obtaining incoherent tokens; thus, truncation of the tail distribution has been heavily discussed, e.g., Top-k \citep{radford2018improving, fan2018hierarchical} and Top-p sampling \citep{holtzman2019curious}. However, a fixed k or p is problematic when considering the high dynamic range of next reasonable tokens, as pointed out in more recent studies on adaptive sampling methods: Mirostat \cite{basu2020mirostat} is proposed based on Zipf statistics and the assumption of a steady perplexity during generation. \citet{hewitt2022truncation} introduce $\eta$-sampling which dismisses the tokens with low probabilities in the tail of the predicted distribution based on absolute and relative thresholds. 
Locally Typical Sampling \cite{meister2023locally} assumes that the generated text should retain a similar entropy rate to that of human-generated text.  Adaptive Decoding \cite{zhu2024improving} proposes to keep the entropy of the truncated distribution close to the original entropy. Although these approaches have been demonstrated to be effective, their performance is highly dependent on the curated truncation parameters and the limited exemplar text.

\subsection{Evaluation of Sampling-based Decoding}
\label{sec:eval}

\noindent \textbf{Common benchmarks} include story generation with WritingPrompts dataset \cite{fan2018hierarchical}, document continuation with WikiText-103 dataset \cite{merity2016pointer} and abstractive summarization on the CNN/DAILYMAIL dataset \cite{nallapati2016abstractive}. These benchmarks suffer from the problem of limited exemplar text, which fails to capture the diverse nature of human language. 

\noindent \textbf{Statistical metrics} are mostly based on n-gram statistics and focus on a single aspect, such as Repetition \cite{welleck2019neural}, Diversity \cite{meister2023locally}, Semantic coherence \cite{gao2021simcse}, Zipf's coefficient \cite{holtzman2019curious} (Unigram rank-frequency) and Self-BLEU \cite{zhu2018texygen}. 

\noindent \textbf{Exemplar-based metrics} dominate the evaluation of sampling-based decoding methods. As observed by \citet{fan2018hierarchical, holtzman2019curious}, lower perplexity of the generated text does not necessarily indicate better quality. And \citet{holtzman2019curious} suggested that the perplexity of the generated text should be close to that of the human text. MAUVE \cite{pillutla2021mauve} takes the trade-off between precision and recall into account, by comparing the learnt distribution from a text generation model to the distribution of human-written text using divergence frontiers. \citet{shi2024thorough} provides a comprehensive evaluation on a large collection of tasks, mostly relying on exemplar-based metrics. However, we reveal that such evaluation is affected by the biases in the curated parameters and limited exemplar text, and our evaluation method is shown to alleviate such an issue. 

% resolved
% \mario{the related work section always needs to say what we do different/better/novel. don't leave it up to the reader.}

\section{Revisiting Truncation Sampling}
% resolved
% \mario{it is better to start a whole section with some text and don't leave it empty. it helps that the reader does not get lost. typically the text here should say how it connects to the previous part of the paper (e.g. intro), what it he main idea, how is the section structured.}
We begin by revisiting the formulation of truncation sampling, followed by identifying the unresolved challenges in evaluating truncation sampling methods.
\subsection{Problem Formulation}
% I followed the notation in the related work for this equation
% \mario{I'm confused the the $\mathcal{A}_{x_{<t}}$ notation. it is difficult to see that this is an index. it looks very similar to $\mathcal{A}_{x<t}$, coudln't it be just $\mathcal{A}_{x_{t}}$? or maybe I haven't understood the semantics of it. }

\begin{definition}
    \begin{equation}
        P_{\textit{trunc}}(x_t|\boldsymbol{x}_{<t}) = \begin{cases} P_{\theta}(x_t|\boldsymbol{x}_{<t})/Z_{\boldsymbol{x}_{<t}} & x \in \mathcal{A}_{\boldsymbol{x}_{<t}} \\
        0 & \mathrm{o.w.},
        \end{cases}
    \end{equation}
    where $\mathcal{A}_{\boldsymbol{x}_{<t}} \, \in \, \mathcal{V}$ denotes the allowed set of candidate next tokens at the $t^{\text{th}}$ position, given a sequence of tokens $\boldsymbol{x}_{<t}=\{x_0, ..., x_{t-1}\}$ as prefix. $Z_{\boldsymbol{x}_{<t}} = \sum_{x \in \mathcal{A}_{\boldsymbol{x}_{<t}}} P_{\theta}(x_t|\boldsymbol{x}_{<t})$ is the renormalization term.
    % resolved
    %\mario{you don't explain $t$ here. }
\end{definition}

Given the Context-Preserving Trie of a reference dataset, we can compute the estimate of the optimal allowed set as follows
% , e.g., the s which exactly covers the full data support Top-k
:   
\begin{definition}
 Let $\mathcal{A}_{\boldsymbol{x}_{<t, \theta}}$ be the allowed set after truncation given the prefix $\boldsymbol{x}_{<t}$. The \textbf{approximated optimal allowed set} $\mathcal{A}^*_{\boldsymbol{x}_{<t}}$ corresponds to the allowed set with the minimum size, while covering the full data support for the $t^{\text{th}}$ 
 % resolved
 % \mario{also here no text in math mode. needs to be e.g. $t^{\textit{th}}$} 
 token $\mathcal{D}_{\boldsymbol{x}_{<t}}$ based on the Trie. It is the solution to the following objective function:
\begin{align}
\begin{split}
       & \mathcal{A}^*_{\boldsymbol{x}_{<t}} = \underset{\theta}{\text{min}}  |\mathcal{A}_{\boldsymbol{x}_{<t, \theta}}| \\
       &\mathrm{s.t.} \quad  \mathcal{D}_{\boldsymbol{x}_{<t}} \subseteq \mathcal{A}_{\boldsymbol{x}_{<t, \theta}}.
\end{split}
\end{align}
\end{definition}
Note that the above definition is designed to exclude the risk of obtaining OOD tokens before the cutoff \cite{finlayson2023closing}, because such type of risk is unsolvable by truncation and is rather determined by the capacity of the trained LLMs. However, such risk is less severe compared to that introduced by inappropriate truncation, since LLMs exhibit a significant capability in predicting the next token \cite{touvron2023llama, achiam2023gpt, jiang2023mistral, team2023gemini} and most OOD samples reside in the tail distribution. 

% \subsection{Parameter-Independent Evaluation}
\subsection{Remaining Issues}
\label{sec:issue}
We reveal three major issues in the evaluation of truncation sampling.  We first summarize the problem of directly using probability as quality metric, then show that the choice of truncation parameter has a significant impact on the evaluation. 

\noindent \textbf{Unreliable Probability}
The probabilities of both the predicted and empirical distribution are not reliable for reflecting the quality of a text.

\begin{itemize}
    \item Higher likelihood does not necessarily imply higher quality of the generated text \cite{fan2018hierarchical, holtzman2019curious, nandwani2023pointwise, wang2024chain}.
     \item Word frequencies are average statistics across various topics, and the optimal probabilities or ranking of each next token is ill-posed. 
    \item  Empirical distribution suffers from the sparsity issue \cite{shareghi2019show, li2016weighted, jurafsky2000speech} of the N-gram models.

    % (show histogram of two datasets, or at least wikipedia vs llm)
    
    % \item Higher frequency of a word does not indicates higher quality. (show an example, a sentence, He is watching the \textit{Film/Movie/Show}, many possibilities)
\end{itemize}

\noindent \textbf{Parameter Sensitivity}
We highlight the complexity and biases in parameter selection: Top-k and Top-p have constant upper bounds, i.e., the vocabulary size $|\mathcal{V}|$ and 1, respectively. In contrast, the upper bounds of $\eta$-sampling and adaptive sampling are dependent on LLM's predicted distribution, because they truncate the tail distribution based on the likelihood of tokens and the slope of Min-Max scaled entropy, respectively. The importance of identifying the effective ranges of such parameters is also reflected in the authors' choice of numeral digit for their parameters. For example, $\Delta\text{Conf}$ is set to $0.0005$ in \citet{zhu2024improving} and $\epsilon$ is chosen from $0.0001$, $0.0009$ and etc in \citet{hewitt2022truncation}. In comparison, the adopted $p$ values for Top-p sampling are merely two digits after zero, such as $0.95$. This shows the significance of identifying the sweet spots of different sampling methods.

\section{Method}
In this section, we derive our metrics for evaluating different sampling-based decoding strategies. The metrics are carefully designed to address the issues discussed in \Cref{sec:issue}.

% The above discussed butterfly effect also suggest that the impact of the tokens with low likelihoods should not be underestimated.

\subsection{Probability-Independent Metrics}
\label{sec:metric}
To circumvent the \textbf{unreliable probability} issue, we merely check whether the predicted next token is in or out of the data support. Specifically, we define \textbf{Recall} and \textbf{Risk} to quantify diversity and quality of a sampling method on a single node of CP-Trie:
\begin{definition} 
\begin{align}
    &\text{Recall}_{\theta, t} = \text{Minimum}\left(\frac{|\mathcal{A}_{\boldsymbol{x}_{<t, \theta}}|}{|\mathcal{A}^*_{\boldsymbol{x}_{<t}}|}, 1\right) \\
    &\text{Risk}_{\theta, t} = \text{Maximum}\left(\frac{|\mathcal{A}_{\boldsymbol{x}_{<t, \theta}}|}{|\mathcal{A}^*_{\boldsymbol{x}_{<t}}|}-1, 0\right)
\end{align}
\end{definition}
$\mathcal{A}_{\boldsymbol{x}_{<t}, \theta}$ is dependent on the parameter selection for truncation, e.g., k value in Top-k sampling. When the allowed set is smaller than the approximated optimal allowed set after truncation, Recall is smaller than one and Risk is regarded as zero. 
With further increased size of the allowed set, Recall reaches one but Risk emerges. Since the sizes of reasonable sets vary drastically for different prefixes, it is not possible to always retain the approximated optimal allowed set with a predefined parameter. In this case, we reveal that the adaptability w.r.t. the varying size of data support of a sampling method indeed determines its effectiveness in real-world application. 

More importantly, our evaluation does not rely on the empirical probability, which is biased and inaccurate due to limited dataset size or context window size. However, the tokens which appear in the dataset could be confidently regarded as reasonable, regardless of their actual probabilities. In addition, considering that temperature could change the flatness of distribution arbitrarily, we adopt ratio of token counts instead of probability mass to make the evaluation independent of temperature tuning and exemplar text. For a detailed discussion with supporting examples, please refer to \Cref{appendix:a}.
% this definition has a problem, because D is usually sparse, i.e., does not cover many Top-k tokens, we simply use the D to determin the largest k and then regard all Top-k tokens as reasonable, then we can obtain a continuous curve
% \begin{definition}
% \begin{equation}
%     \textit{Recall} = \frac{P(\mathcal{A}_{\boldsymbol{x}_{<t, \theta}} \bigcap \mathcal{D}_{\boldsymbol{x}_{<t}})}{P(\mathcal{D}_{\boldsymbol{x}_{<t}})},
% \end{equation}
% where $\mathcal{A}_{\boldsymbol{x}_{<t}, \theta}$ is the allowed set of a sampling method, and $\mathcal{D}_{\boldsymbol{x}_{<t, \theta}}$ denotes the data support given a prefix $\boldsymbol{x}_{<t}$.
% \end{definition}

% Based on the above analysis, we propose to only compare the Recall of different sampling methods within a low-risk region. 

% and adopt $\textit{Recall}_{\textit{Risk-1}}$, $\textit{Recall}_{\textit{Risk-5}}$ and $\textit{Recall}_{\textit{Risk-10}}$ for quantitative evaluation.
\subsection{Tuning-Independent Evaluation}
\label{sec:param}
To eliminate the huge impact of \textbf{Parameter Sensitivity} issue on fair evaluation, we adopt  \textbf{Average Recall (AR)} at an average Risk and \textbf{Risk Standard Error (RSE)} at an average Risk to quantify \textbf{diversity} and \textbf{stability}  of a sampling method across 
$N$ nodes of CP-Trie, respectively: 
\begin{definition}
\begin{equation}
\begin{split}
 &\text{AR}_{\text{Risk}-0.1} = \frac{1}{N}\sum_{i=1}^N\text{Recall}^{(i)}_{\theta,t} \\ 
  & \resizebox{0.48\textwidth}{!}{$\text{RSE}_{\text{Risk}-0.1} = \frac{1}{N}\sqrt{\sum_{i=1}^N(\text{Risk}^{(i)}_{\theta,t}-\frac{1}{N}\sum_{i=1}^N\text{Risk}^{(i)}_{\theta,t})^2}$}\\
  &\mathrm{s.t.} \quad \frac{1}{N}\sum_{i=1}^N\text{Risk}^{(i)}_{\theta, t} = 0.1,
 \end{split}
\end{equation}
where the superscript $(i)$ denotes the $i^{th}$ node in the evaluation set of nodes on the prefix tree.
Analogously, a family of critical values such as $\text{AR}_{\text{Risk}-0.5}$ can be easily defined. 
\end{definition}

% \begin{definition}
% \begin{equation}
% \begin{split}
%  & \resizebox{0.48\textwidth}{!}{$\text{RSE}_{\text{Risk}-\textit{0.1}} = \frac{1}{N}\sqrt{\sum_{i=1}^N(\text{Risk}^{(i)}_{\theta,t}-\frac{1}{N}\sum_{i=1}^N\text{Risk}^{(i)}_{\theta,t})^2}$} \\ &s.t. \quad \frac{1}{N}\sum_{i=1}^N\text{Risk}^{(i)}_{\theta, t} = 0.1.
%  \end{split}
% \end{equation}
% \end{definition}

Since $\theta$ is now determined by the given average Risk, the diversity metric reflects the genuine capacity of a sampling method regardless of parameter tuning. This allows for a fair comparison of different sampling methods, especially considering their drastically different effective ranges, as mentioned in \Cref{sec:intro} and \Cref{sec:issue}.

% \subsection{The Priority of Low Variance in Risk }
% \label{sec:low-variance}

% \begin{conjecture}
% At a given average risk level, the total amount of risk when generating a sequence of length $T$ is reduced with decreased variance of the risks at each decoding step.  
% \end{conjecture}

% The above conjecture can be mainly understood by the auto-regressive generation process of LLMs, as shown in \Cref{eq:butter}.
% % \begin{equation}
% %  P_{\theta}(\exists x_t \not\in \mathcal{A}^*_{\boldsymbol{x}_{<t}}) = 1 - \prod_{t=1}^{T}P_{\theta}({x_{t} \in \mathcal{A}^*_{\boldsymbol{x}_{<t}}|\boldsymbol{x}_{<t}}).   
% % \end{equation}
% The in-distribution probability dependent on the risk at each decoding step and is minimized when the product of the probabilities is maximized at a given average risk level. We could infer that the sum of the in-distribution probabilities is approximated determined at a given average risk level. For simplification, we assume their sum is unchanged. According to AM-GM inequality, the maximum of the product is achieved when each individual component is equal to each other. Roughly speaking, the less variance of the probability masses at each step, the larger their product and thus the smaller the total risk is.  

\section{Experiment}

\begin{figure}[ht]
    \centering
    \includegraphics[width=0.48\textwidth]{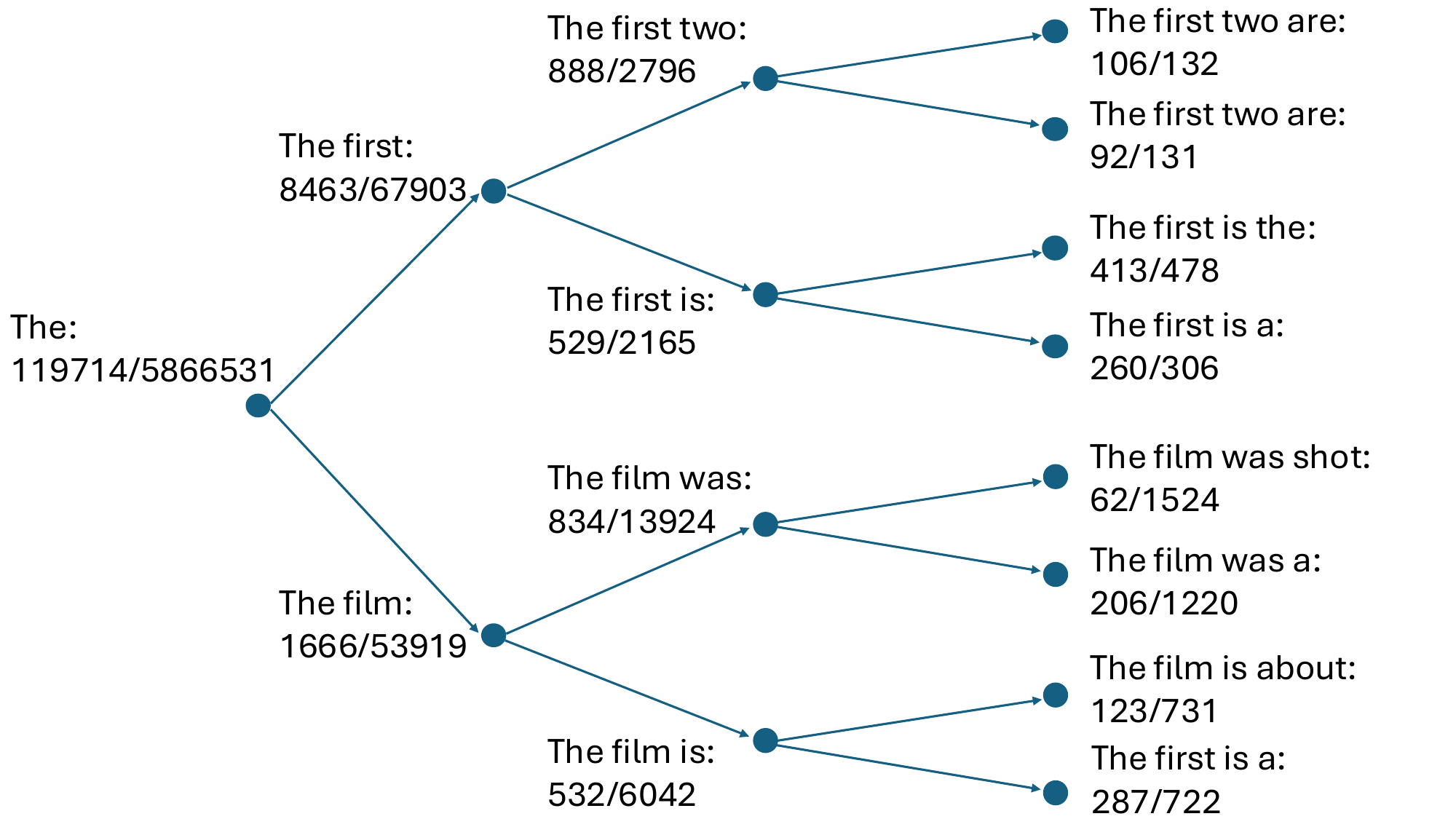}
    \caption{Illustration of the EnWiki CP-Trie. For brevity, only two child nodes are shown at each depth. The number at the left side of the slash symbol refers to the branching factor at the current node, and the number at the right side refers to the total number of leaves of the sub-tree with the current node as the root node.}
    \label{fig:tree}
\end{figure}

In this section, we conduct evaluation of existing sampling-based decoding approaches on our collected EnWiki CP-Trie dataset. We aim to estimate the inherent adaptability of sampling-based methods and the results could be used as references for the application of LLMs in open-ended tasks.

% \mario{also state the goals of the experimental section here. which hypothesis are going to be evaluated/tested.}

\subsection{Data Collection}
We construct our Trie data based on the English subset of Wikipedia dataset, named EnWiki CP-Trie. As shown in \Cref{fig:tree}, all possible words that appear after a given prefix in the dataset are treated as child nodes, with their preceding word regarded as the parent node. Starting from "Begin of Sequence" and collecting the child nodes recursively, we are able to transform the full dataset into a single prefix tree. We elaborate the main design choices in the following:

\noindent \textbf{Basic Unit.} 
% resolved
% \mario{I have a preference to put a "." after these paragraph headings -- like \textbf{Basic Unit.} But up to you.} 
It is possible to split the datasets into articles, paragraphs, sentences or n-grams. Constructing a tree based on articles or paragraphs may require more data than the training data of LLMs to guarantee an adequate number of branches (because LLMs lean to interpolate), whereas the construction based on n-grams suffers from poor contextual information and is heavily biased towards common tuplets of n tokens regardless of the context. Therefore, we adopt sentence as the basic unit, which guarantees a coherent context at sentence-level and requires much fewer data than training. It is noteworthy that a n-gram Trie \cite{jurafsky2000speech} tends to overestimate the data support size given a prefix \cite{bengio2000neural}, due to the loss of information outside the contextual window, as shown in \Cref{fig:example}.

\begin{figure}[h]
    \centering
\includegraphics[width=0.43\textwidth]{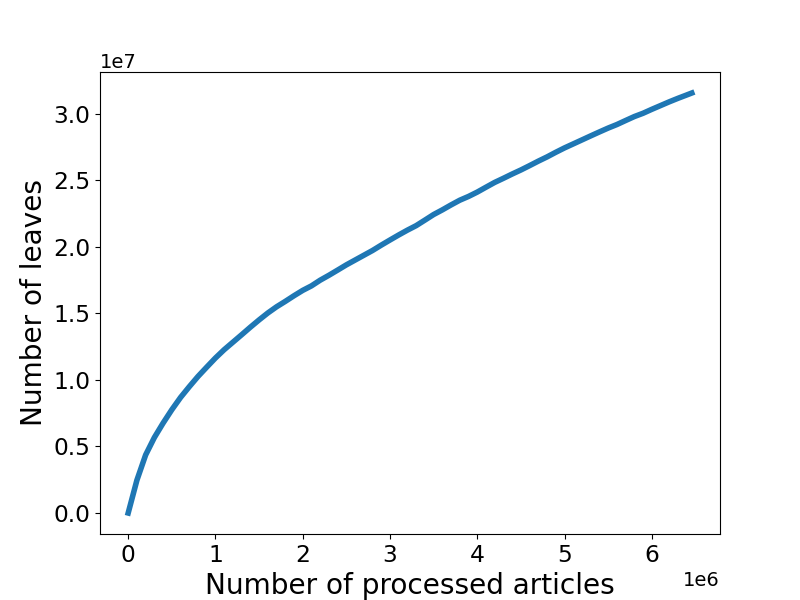}
    \caption{The total number of leaves on the CP-Trie against the total number of processed articles. 
    % resolved
    %\mario{this plot looks very sketchy and not publication ready. larger font size, thicker line.}
    }
    \label{fig:stats}
\end{figure}

\noindent \textbf{Filtering.} To avoid invalid words or rare proper names which are unreasonable for the model to predict, we exclude the sentences containing such words by checking their presence in the WORD LIST dataset, which is available on the website \footnote{\href{https://web.archive.org/web/20131118073324/https://www.infochimps.com/datasets/word-list-350000-simple-english-words-excel-readable}{word-list dataset homepage}}. It contains 354986 words in total and explicitly excludes proper names and compound words. Section titles are also excluded, which are often incomplete sentences with poor contextual information.

\noindent \textbf{Statistics.} Wikipedia-English dataset contains $6,458,670$ articles, which result in EnWiki CP-Trie with $31,557,359$ leaves, see \Cref{fig:stats}.

% also check the number of sentences after filtering, comparing to the number of leaves

\noindent \textbf{Storage.} The prefix tree is implemented as a nested dictionary and saved in JSON format. Since each lookup at any depth has constant complexity, the retrieval is highly efficient. Moreover, the dictionary is easily extendable if extra data are needed for a more accurate estimation of the full data support.

\subsection{Evaluation Setup}

\noindent \textbf{Baselines.}
Our evaluation includes Top-k sampling \cite{radford2018improving,fan2018hierarchical}, Top-p sampling \cite{holtzman2019curious}, $\eta$-sampling \cite{hewitt2022truncation}, Adaptive sampling \cite{zhu2024improving} and Mirostat \cite{basu2020mirostat} into comparison.

% \noindent \textbf{Evaluation Protocol}
\noindent \textbf{Evaluation Data.} To guarantee a tight lower bound of the ideal data support given different prefixes, we first sort the sub-nodes according to their total number of leaves at each depth, then we select the top 10 sub-trees with different sentence starting tokens for evaluation. Moreover, we keep the top 2 child nodes at each depth till depth 6, since the empirical data support becomes less adequate at large depth. This results in an evaluation set of $593$ prefixes with varying lengths in total.

\noindent \textbf{Evaluation Metrics.} We measure the improvement in \textbf{diversity} via the increase of \textbf{Average Recall(AR)} at an average Risk, and the improvement of the \textbf{stability} at each decoding step in the auto-regressive process via the decrease of \textbf{Risk Standard Error (RSE)} at an average Risk. We adopt \textbf{AR} and \textbf{RSE} at average Risks of 1, 5 and 15 for comparison, representing low, medium, and high-risk regions, respectively. 

\noindent \textbf{LLMs.}
To ensure that the conclusion generalizes to different models, we adopt Llama \cite{touvron2023llama, dubey2024llama} family, Mistral \cite{jiang2023mistral, jiang2024mixtral} 
family  and GPT-2-XL \cite{radford2019language} for comparison. 

\noindent \textbf{Tokenization.} Since different LLMs are trained with different encoding methods, the evaluation has to be independent of the encoding methods. We solve this issue by constructing the CP-Trie with either a word or punctuation. For example, if the predicted next token corresponds to ``sec", which is a part of the in-distribution word "section", then we regard this as a correct prediction. The second part ``tion" is regarded as a hidden child node and is skipped in the evaluation.

\noindent \textbf{Parameter Search.} We apply grid search to determine the corresponding parameters of different sampling methods for each average Risk. To address the highly non-linear dependency between the sampling methods and their truncation parameters, we employ an efficient coarse-to-fine grid search strategy: the number of grids is initially set to 2000. If a parameter results in an average Risk within ±0.1 of the target value, it is considered a feasible solution. Otherwise, an additional grid search is performed within a smaller interval until a feasible solution is found, based on the initial search results. The grids are determined using Llama3-70B and are applied consistently across all models. As shown in \Cref{tab:detail1}, almost all the deviations in the average Risks are much smaller than 0.1, demonstrating the robustness of our strategy. 

\noindent \textbf{Implementation.}
Our implementation mainly relies on Pytorch \cite{paszke2017automatic}, HuggingFace \cite{wolf2019huggingface} and OpenAI API \footnote{\url{https://pypi.org/project/openai/}} library. We implement a truncation sampling method ourselves if the official implementation is unavailable. For all methods, the minimum size of the allowed set is set to 1 to prevent breaking the sampling process.

\begin{table*}[ht]
    \centering
    % \footnotesizeƒt
    \scriptsize
    \begin{tabular}{@{}c@{\hspace{0.2cm}}|@{\hspace{0.2cm}}c@{\hspace{0.2cm}}|@{\hspace{0.2cm}}c@{\hspace{0.2cm}}|@{\hspace{0.2cm}}c@{\hspace{0.2cm}}c@{\hspace{0.2cm}}|@{\hspace{0.2cm}}c@{\hspace{0.2cm}}|@{\hspace{0.2cm}}c@{\hspace{0.2cm}}c@{\hspace{0.2cm}}|@{\hspace{0.2cm}}c@{\hspace{0.2cm}}|@{\hspace{0.2cm}}c@{\hspace{0.2cm}}c@{}}
    \toprule
  \multirow{2}*{Model} &\multirow{2}*{Method} &
  \multicolumn{3}{@{\hspace{0.1cm}}c@{\hspace{-0.5cm}}}{Avg. Risk 1}&
  \multicolumn{3}{@{\hspace{0.1cm}}c@{\hspace{0.1cm}}}{Avg. Risk 5}&
  \multicolumn{3}{@{\hspace{0.1cm}}c@{\hspace{0.1cm}}}{Avg. Risk 15}\\
    % \midrule
    \cmidrule{3-11}
     &  & Parameter & RSE $\downarrow$  & AR $\uparrow$
 & Parameter & RSE $\downarrow$ & AR $\uparrow$& Parameter & RSE $\downarrow$ & AR $\uparrow$\\
       % \midrule
        % \multicolumn{10}{c}{GPT2-XL} \\
              \midrule
  \multirow{5}*{\rotatebox{90}{GPT2-XL}}    

        &Adaptive  & 9.5e-4 &  0.006 & \textbf{0.252}  &1.1e-4& 0.679 & \textbf{0.339}& 2.5e-05 & 2.241 & \textbf{0.413}  \\
              &Mirostat &   4.425 & \textbf{0.005} & 0.236  & 5.9475& 0.717 & 0.326 & 6.76 & 2.501 & 0.401 \\
              &  Top-k &   15    & 0.006 & 0.220 & 64  & \textbf{0.613} & 0.290  &    184 & \textbf{1.781} & 0.340 
   \\
        &Eta  &  0.318 & 0.013&0.198  & 0.011& 1.484 & 0.301 &  0.001 & 4.261 & 0.404  \\
         & Top-p &    0.5705 &\underline{0.015} & \underline{0.170}    &0.746& \underline{2.129} & \underline{0.240}  &  0.8555   & \underline{6.210}  & \underline{0.338}\\

\midrule
      % \multicolumn{10}{c}{Llama-2-7b} \\
          % & Parameter & Risk & Recall  & Parameter & Risk & Recall & Parameter & Risk & Recall \\
              % \midrule
  \multirow{5}*{\rotatebox{90}{Llama-2-7b}} 

    &Adaptive &   1.1e-3 & 0.154 & \textbf{0.257} & 1.4e-4 & 0.856 & \textbf{0.364}  & 3.1e-5 & 2.966 & 0.470 
    \\
        &Mirostat &  4.253 & 0.133 & 0.236  & 5.82 & 0.650 & 0.349  & 6.628 & 2.286 & \textbf{0.474} \\
        &  Top-k & 14 & \textbf{0.126} & 0.226  & 61 & \textbf{0.587} & 0.296  & 177 & \textbf{1.722} & \underline{0.369} 
    \\ 
    &Eta &  0.512 & \underline{0.563}  & 0.192   & 0.023 & \underline{2.599} & 0.297 & 0.002 & \underline{6.531} & 0.407 
    \\
        &Top-p & 0.54 & 0.529 & \underline{0.156}  & 0.7665 & 2.331 & \underline{0.254}  & 0.9 & 6.208 & 0.400 
    \\

    % \midrule
  % \multicolumn{10}{c}{Llama-3-8B} \\
          % & Parameter & Risk SE & Recall Mean  & Parameter & Risk & Recall & Parameter & Risk & Recall \\
              % \midrule
                      \midrule
   \multirow{5}*{\rotatebox{90}{Llama-2-70b}}     
        &Adaptive & 0.0011 &    0.142 &     \textbf{ 0.269}  & 1.2e-4&          0.796 &      0.374 
 & 2.3e-5 &     2.697 &      0.485 
 \\
        &Mirostat & 4.16 & 0.135 &      0.238 
 & 5.7875 &        0.684 &      0.353 
 &6.67 &    2.125 &      0.478 
 \\
 &Top-k & 14 & \textbf{ 0.128} &      0.232  & 60 &      \textbf{0.583} &      0.307 
& 174 &     \textbf{1.712} &      \underline{0.375} 
\\
        &Eta & 0.092 &   0.304 &      0.236 & 0.003 &      1.590 &     \textbf{ 0.378} 
& 2.1e-4&      4.243 &      \textbf{0.510} 
\\

    &Top-p & 0.6535 &  \underline{0.475} &      \underline{0.189} & 0.8465 &        \underline{2.136} &      0.316 
& 0.9395 &  \underline{5.522} &      0.468 
\\
 \midrule
    \multirow{5}*{\rotatebox{90}{Llama-3-8B}}  
    &Adaptive &  1.1e-3 &  0.167 & \textbf{0.260} & 1.7e-4& 0.787 & \textbf{0.343}  & 3.7e-5 &       2.685 &     \textbf{ 0.418} \\
    &Mirostat &  4.24 & 0.139 & 0.230  & 5.8175 & 0.804 & 0.318 & 6.693 &          2.630 &      0.393 \\
     &Top-k &  14 & \textbf{0.128} & 0.228  & 59 & \textbf{0.576} & 0.290 & 172 &       \textbf{ 1.701} &      0.346 \\
         &Eta &  0.673 & 0.445 & 0.181  & 0.029 & \underline{2.112} & 0.271 & 0.002 &       \underline{6.009} &      0.373\\
             &Top-p &   0.5395 &  \underline{0.451} & \underline{0.154}  &0.736 &  2.061 & \underline{0.224}  & 0.855 &     5.770 &      \underline{0.326} \\

    % \midrule
      % \multicolumn{10}{c}{Llama-3-70B} \\
 
          \midrule
    \multirow{5}*{\rotatebox{90}{Llama-3-70B}}  
    &Adaptive &   1.1e-3 &      0.137 &      \textbf{0.263}  &  1.4e-4&         0.787 &      \textbf{0.353} &3.16e-5&    2.778 &      \textbf{0.424}
 \\
     &Mirostat &  4.21 & 0.138 &      0.230  &5.91 &        0.708 &      0.332 &6.84 &    2.193 &      0.417 \\
 &Top-k & 14 & \textbf{0.127} &      0.230 & 60&      \textbf{0.581} &      0.295    & 173 &      \textbf{1.695} &      0.352 \\
    &Eta &   0.37 &  0.137 &      \textbf{0.263}  & 0.014&          2.231 &      0.295 &0.001  &         6.265 &      0.398 \\
  &Top-p &  0.5695 &     \underline{0.502} &     \underline{0.158} &   0.758  &         \underline{2.386} &      \underline{0.237} &0.8705&     \underline{6.685} &      \underline{0.332} \\

%  \multirow{2}*{Method}   & \multicolumn{9}{c}{Llama-2-70b} \\
        % &  & Parameter & Risk & Recall  & Parameter & Risk & Recall & Parameter & Risk & Recall \\
        %   \hline

%         \hline

\midrule
   \multirow{5}*{\rotatebox{90}{Mixtral-7B}}

      &  Adaptive & 0.00105  &      0.152 &      \textbf{0.260} & 1.2e-4 &          0.809 &      0.364 
        & 2.2e-5  &    2.757 &      0.466 
        \\
             & Mirostat &4.1825 &     0.141 &      0.236  & 5.8125 &          0.721 &      0.345  
        & 6.71 &    2.213 &      0.468 
        \\
        & Top-k &  14  &          \textbf{ 0.126} &      0.224 & 62 &     \textbf{0.596} &     \underline{0.297} 
        & 181 &     \textbf{1.759} &     \underline{0.364}
        \\
       & Eta & 0.075 &     0.307 &      0.243  & 0.003 &     1.542 &      \textbf{0.368} 
        & 1.96e-4&    4.712 &     \textbf{ 0.505} 
        \\
         &   Top-p & 0.6565 &  \underline{0.539} &      \underline{0.194} & 0.8375 &     \underline{2.476} &      0.303 
& 0.9315 &    \underline{6.315} &      0.447 \\
 
      \midrule
   \multirow{5}*{\rotatebox{90}{Mixtral-8x7B}} 
       & Adaptive &  0.00105 &  0.148 &     \textbf{ 0.265}  & 1.1e-4&        0.798 &      0.372 
 & 2.1e-5 &   2.802 &      0.476 
 \\
        & Mirostat & 4.2775 &   0.143 &      0.238  & 5.845 &         0.710 &      0.346 
        & 6.6875 &    2.213 &      0.461 
        \\
    &  Top-k & 15 &     \textbf{ 0.134} &      0.229  & 63 &     \textbf{0.598} &      \underline{0.301} 
        & 183   &     \textbf{1.757} &      \underline{0.366} 
        \\
       & Eta & 0.087 &    0.335 &      0.241  & 0.003  &     1.822 &     \textbf{ 0.375} 
& 2.15e-4&    4.922 &      \textbf{0.506}
\\
           & Top-p & 0.6505 &   \underline{0.535} &      \underline{0.192}  & 0.8375 &         \underline{2.423} &      0.303 
        & 0.9325 &    \underline{6.139} &      0.456 
        \\

%         \hline
   
%     \multirow{2}*{Method}   & \multicolumn{9}{c}{Mistral-7B} \\
%           & Parameter & Risk & Recall  & Parameter & Risk & Recall & Parameter & Risk & Recall \\
%           \hline
%         Top-k &  14  &      0.965 (     0.126) &      0.224 (     0.016) & 62 &      4.968 (     0.596) &      0.297 (     0.017)
%         & 181 &     15.006 (     1.759) &      0.364 (     0.018)
%         \\
%         Top-p & 0.6565 &  1.001 (     0.539) &      0.194 (     0.014) & 0.8375 &      4.996 (     2.476) &      0.303 (     0.016
% ) 
% & 0.9315 &     15.038 (     6.315) &      0.447 (     0.016)
% \\
%         Adaptive & 0.00105  &      1.001 (     0.152) &      0.260 (     0.016) & 0.000115 &      4.993 (     0.809) &      0.364 (     0.018)
%         & 2.2e-5  &     14.999 (     2.757) &      0.466 (     0.017)
%         \\
%         Eta & 0.075 &      0.997 (     0.307) &      0.243 (     0.015) & 0.003 &      4.640 (     1.542) &      0.368 (     0.017) 
%         & 0.000196 &     15.009 (     4.712) &      0.505 (     0.017)
%         \\
%         Mirostat &4.1825 &      1.000 (     0.141) &      0.236 (     0.016) & 5.8125 &      4.999 (     0.721) &      0.345 (     0.018) 
%         & 6.71 &    14.978 (     2.213) &      0.468 (     0.018)
%         \\
        
        \bottomrule
    \end{tabular}
    \caption{Risk Standard Error (RSE, indicating stability) and Average Recall (AR, indicating diversity) of different truncation sampling methods at different average Risks using different models. The corresponding parameter of each method at an average risk level is also provided. The best and worst scores are marked in bold and underlined, respectively. For more detailed results, please refer to \Cref{appendix:b}.}
    \label{tab:reference}
\end{table*}

\subsection{Comparison at Different Average Risks}
In this section, we conduct a comprehensive study of different truncation sampling methods at different average Risks. As discussed in \Cref{sec:param}, this allows for a fair comparison which is independent of parameter tuning. Moreover, we provide the corresponding parameters for each truncation sampling method at different average Risks, which could serve as a user reference for these methods.

As can be seen in \Cref{tab:reference}, different truncation sampling methods are compared at the average Risk of 1, 5, and 15 respectively. As discussed in \Cref{sec:metric}, our defined risk and recall metrics explicitly exclude the source of risk induced by a LLM's capacity by design, thus similar parameter values correspond to the same risk level for most sampling methods across various model types and sizes. This exactly showcases the advantage of our evaluation being tuning-independent and sustainable to the rapid update of LLMs. Among the evaluated methods, Eta-sampling \cite{hewitt2022truncation} is sensitive to the changes of model type and size, which might hinder its practical significance especially at a low risk level. 

Regarding diversity, i.e., the average recall at the same average Risk, Adaptive sampling \cite{zhu2024improving} and Mirostat \cite{basu2020mirostat} are the best and second performers, which consistently outperform the Top-k baseline by a considerable margin. Top-p mostly exhibits inferior recall comparing to the Top-k baseline, so does Eta-sampling at the average Risk of 1. As for the stability represented by standard error of Risks, Top-k sampling reaches the best scores in most cases. In comparison, Adaptive sampling and Mirostat deliver comparable standard error of risks to Top-k sampling, whereas Top-p sampling and Eta-sampling are again inferior. Considering both diversity and stability, Adaptive sampling and Mirostat are the top 2 adaptive methods to be recommended, whereas Top-p sampling shall be the last two methods to be considered. 

\begin{figure}[ht]
    \centering
    \begin{subfigure}[t]{0.15\textwidth}
    \centering
    \includegraphics[width=\textwidth]{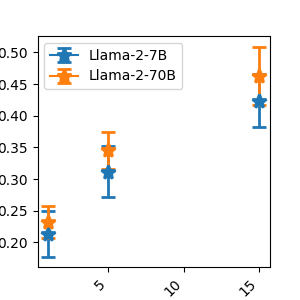}
    \caption{Llama-2 family.}    
    \end{subfigure}
     \begin{subfigure}[t]{0.15\textwidth}
    \centering
    \includegraphics[width=\textwidth]{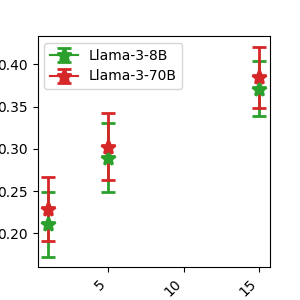}
    \caption{Llama-3 family.}    
    \end{subfigure}
         \begin{subfigure}[t]{0.15\textwidth}
    \centering
    \includegraphics[width=\textwidth]{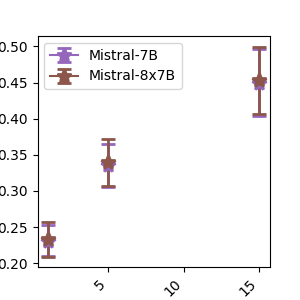}
    \caption{Mistral family}    
    \end{subfigure}
    \caption{Comparing the average Recalls at given average Risks using different model sizes.}
    \label{fig:size}
\end{figure}

We also show in \Cref{fig:size} that larger models of the same family have higher average recall at the same risk level comparing to the smaller ones. This conforms to the fact that larger models better captures the human text distribution. Please note that our metrics does not allow a direct comparison between different model families, mainly due to their different vocabulary sizes and tokenizers, e.g., Llama-3 has a 128,256 vocabulary size, while Llama-2 has only 32, 000 vocabulary size. Moreover, our metrics also explicitly exclude the source of risk within the optimal allowed set, which is heavily dependent on a LLM's capacity.

% \begin{figure}[ht]
%     \centering
%     %\frame
%     {
%     \scriptsize%\url
%     \begin{tabular}{@{}p{0.47\columnwidth}p{0.47\columnwidth}@{}}
%     \toprule
%     Question&gpt2xl\\
%     \midrule
%     What happens to you if you eat watermelon seeds?& You will die.\vspace{0.4cm}\\
%     Where did fortune cookies originate?& Fortune cookies originated in the United States.\\
%     \bottomrule
% %    \includegraphics[width=\linewidth]{Screenshot 2024-08-15 at 16.08.15.png}
%     \end{tabular}
%     }
%     \caption{Greedy decoding might exclude the paths of correct answers, as shown in the answer to the first question taken from the question list of TruthfulQA.}
%     \label{fig:example-greedy}
% \end{figure}

\begin{table}[ht]
    \centering
    \scriptsize
    \label{tab:acc}
    \begin{tabular}{@{}lccc@{}}
      \toprule
    \multirow{2}*{Methods} &     \multicolumn{3}{c}{ Mean(std) Accuracy $\uparrow$ }
     \\
     \cmidrule{2-4}
    &  Avg. Risk 1 & Avg. Risk 5 & Avg. Risk 15 \\
     \midrule
    Greedy &  \multicolumn{3}{c}{0.338}  
    \\
    \midrule
    Na\"ive & \multicolumn{3}{c}{0.421(0.004)} \\
    \midrule
    Top-k  & 0.401(0.010)& \textbf{0.436}(0.008)&0.421(0.010)\\
    Top-p  
    & \underline{0.355}(0.013) & \underline{0.378}(0.011) & \underline{0.389}(0.012)
    \\
    Adaptive  
    & 0.395(0.012) & 0.424(0.011) & 0.421(0.009)
    \\
    Eta  
    & 0.388(0.005) & 0.401(0.013) & 0.413(0.026)
    \\
    Mirostat 
    & 0.413(0.010) & 0.425(0.013) & \textbf{0.425}(0.009)
    \\
    % Typical &  \textbf{0.427}(0.020) & 0.411(0.025) & 0.408(0.011)  \\
   
\bottomrule
    \end{tabular}
       \caption{Evaluation on the TruthfulQA benchmark under the open-ended generation setup. Naive sampling refers to sampling without truncation. The best and worst scores are marked in bold and underlined, respectively. For more details, please refer to \Cref{appendix:b}.}
\end{table}

% \subsection{Qualitative Results}
\begin{figure}[ht]
\captionsetup{font=footnotesize,labelfont=footnotesize}
    \centering
 \begin{subfigure}[t]{0.156\textwidth}
    \centering
    \includegraphics[trim={0cm 0 0cm 0}, width=\textwidth]{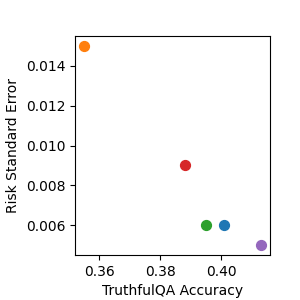}
    \caption{Correlation at Avg. Risk 1: $-0.87$}    
    \end{subfigure}
     \begin{subfigure}[t]{0.156\textwidth}
    \centering
    \includegraphics[trim={0cm 0 0cm 0}, width=\textwidth]{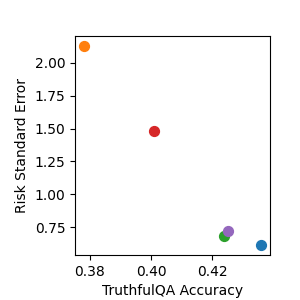}
    \caption{Correlation at Avg. Risk 5: $-0.92$}    
    \end{subfigure}
     \begin{subfigure}[t]{0.156\textwidth}
    \centering
    \includegraphics[trim={0cm 0 0cm 0}, width=\textwidth]{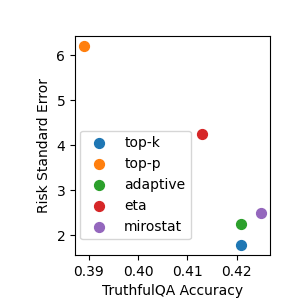}
    \caption{Correlation at Avg. Risk 15: $-0.94$}    
    \end{subfigure}
     \begin{subfigure}[t]{0.156\textwidth}
    \centering
    \includegraphics[trim={0cm 0 0cm 0}, width=\textwidth]{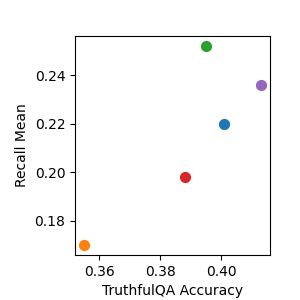}
    \caption{Correlation at Avg. Risk 1: $0.83$}    
    \end{subfigure}
     \begin{subfigure}[t]{0.156\textwidth}
    \centering
    \includegraphics[trim={0cm 0 0cm 0}, width=\textwidth]{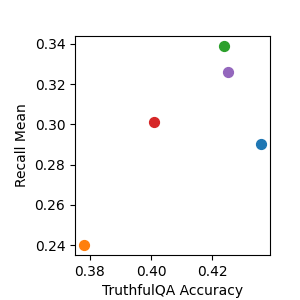}
    \caption{Correlation at Avg. Risk 5: $0.83$}    
    \end{subfigure}
     \begin{subfigure}[t]{0.156\textwidth}
    \centering
    \includegraphics[trim={0cm 0 0cm 0}, width=\textwidth]{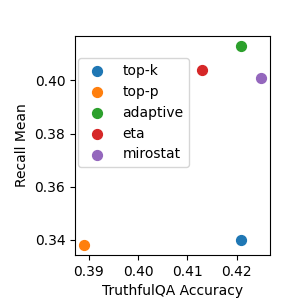}
    \caption{Correlation at Avg. Risk 15: $0.50$}    
    \end{subfigure}
    \caption{The scatter plots of TruthfulQA accuracy against risk standard error (first row) and recall mean (second row) at different average Risks.}
    \label{fig:cor}
\end{figure}

\subsection{Validation on TruthfulQA Benchmark}
Although our evaluation protocol is grounded by the thorough design process with reasonable simplifications, we would like to verify its effectiveness in the real-world scenario using the TruthfulQA Benchmark \cite{lin2021truthfulqa}. The evaluation results using gpt2-xl are shown in \Cref{tab:acc}. For all the methods other than greedy decoding, we run 3 times at each average risk level and report the mean and standard deviation (parenthetical value). 

It can be observed that greedy decoding falls far behind sampling-based decoding strategies, which conforms to the issue of likelihood-oriented decoding discussed in \Cref{sec:intro}, as well as the findings in recent studies \cite{cobbe2021training, wang2023self, wang2024chain, shi2024thorough}. All the truncation sampling methods at the low risk level achieves lower accuracy comparing to Naive sampling, due to the over-truncation of the decoding paths. At the average risk level of 5, all the truncation sampling methods slightly improve their own accuracy. Top-k sampling, Adaptive sampling and Mirostat also reach comparable or slightly higher accuracy in comparison to Naive sampling. However, further increased average risk level (means improved average recall and thus diversity) does not benefit the performance on TruthfulQA, which is plausible. Moreover, there exists a even stronger correlation between Risk SE (Standard Error of Risks) and TruthfulQA accuracy, validating the importance of  stability when evaluating an adaptive decoding method. The strong correlation between TruthfulQA accuracy and our proposed average recall as well as standard error of risks at different average Risks validate the soundness and effectiveness of our evaluation method. 

\section{Revisiting Existing Evaluation}
In this section, we revisit the recent study \citep{shi2024thorough} by comparing sampling-based decoding methods at the same average Risks. We adopt the official implementation of \citet{shi2024thorough}. Following their  setups, we adopt Llama-2-7B on MBPP \citep{austin2021program}, HumanEval \citep{austin2021program} and GSM8K \citep{cobbe2021training} to evaluate coding and math problem solving performance. Mean and standard deviation for three runs are reported in \Cref{tab:mbpp}, \Cref{tab:human_eval} and \Cref{tab:gsm8k}, respectively.

For all the three tasks, Mirostat does not perform well in general, probably because it is based on the Zipf-law of natural language and thus not suitable for code and math tasks. Notably, our greedy decoding baseline achieves significantly lower result than reported by \citet{shi2024thorough} on HumanEval. Our results should be plausible, because the instruction tuned Llama-2-7B only achieves 7.9 according to Meta-Llama Github\footnote{
\resizebox{0.44\textwidth}{!}{
\url{https://github.com/meta-llama/llama3/blob/main/MODEL_CARD.md}}
}
. 

While their study concludes that deterministic methods outperform sampling methods across most tasks, our evaluation reveals that sampling methods are indeed underestimated. In contrast to the conclusion in \citet{shi2024thorough}, all the sampling-based decoding methods could achieve better performance than greedy decoding on HumanEval in \Cref{tab:human_eval}. In addition, Top-p and eta sampling also beat greedy decoding at a low average Risk on GSM8K in \Cref{tab:gsm8k}. This observation underscores the challenges in parameter selection for sampling-based decoding, which is effectively addressed by our method.

\begin{table}[h]
    \centering
    \scriptsize
    \begin{tabular}{c|ccc}
    \toprule
    Methods &Avg. Risk 1 &Avg. Risk 5 &Avg. Risk 15 \\
    \midrule
    Top-k & 19.70 (0.50) & 21.00 (2.30) & 20.50 (0.30) \\
    Top-p & 21.50 (1.30) & \textbf{21.10} (0.40) & \textbf{21.70} (0.70) \\
    Mirostat & \underline{9.50} (0.30) & \underline{8.80} (2.00) & \underline{8.80} (0.40) \\
    Eta & \textbf{22.10} (0.70) & 19.10 (0.40) & 19.70 (0.40) \\
    \hline
    Greedy & \multicolumn{3}{c}{24.00} \\
    \bottomrule
    \end{tabular}
    \caption{Pass@1 accuracy on MBPP. It is consistent to the observation by \citet{shi2024thorough} that sampling methods are inferior to greedy decoding.}
    \label{tab:mbpp}
\end{table}

\begin{table}[h]
    \centering
    \scriptsize
    \begin{tabular}{c|ccc}
    \toprule
    Methods &Avg. Risk 1 &Avg. Risk 5 &Avg. Risk 15 \\
    \midrule
    Top-k & \textbf{5.68} (2.00) & 5.08 (1.52) & 6.50 (0.76) \\
    Top-p & 3.46 (1.05) & 5.89 (0.76) & \textbf{6.52} (2.43) \\
    Mirostat & 3.25 (0.76) & \underline{4.27} (1.00) & \underline{4.68} (0.58) \\
    Eta & \underline{2.64} (2.01) & \textbf{6.91} (1.04) & 6.10 (1.32) \\
    \hline
    Greedy & \multicolumn{3}{c}{2.44} \\
    \bottomrule
    \end{tabular}
    \caption{Pass@1 accuracy on HumanEval. Sampling methods perform better with higher average Recalls and Risks.}
    \label{tab:human_eval}
\end{table}

\begin{table}[h]
    \centering
    \scriptsize
    \begin{tabular}{c|ccc}
    \toprule
   Methods & Avg. Risk 1 &Avg. Risk 5 &Avg. Risk 15\\
   \midrule
   Top-k & 7.56 (5.39) & \textbf{11.90} (0.80) & \textbf{11.73} (0.57) \\
   Top-p & \textbf{14.13} (0.47) & 8.72 (6.18) & 11.67 (0.11) \\
   Mirostat & \underline{5.46} (0.47) & \underline{5.74} (0.64) & \underline{3.46} (2.10) \\
   Eta & 13.72 (0.46) & 8.42 (5.54) & 11.22 (0.75) \\
   \hline
      Greedy & \multicolumn{3}{c}{13.19} \\
      \bottomrule
  \end{tabular}
    \caption{Accuracy on GSM8K. Top-p and eta sampling outperforms greedy decoding at an average Risk of 1.}
    \label{tab:gsm8k}
\end{table}

% \begin{table}[t]
%     \centering
%     \scriptsize
%     \begin{tabular}{c|ccc}
%     Sampling Method & risk 1 & risk 5 & risk 15 \\
%     \hline
%     topk & 0.0568 (0.0200) & 0.0508 (0.0152) & 0.0650 (0.0076) \\
%     topp & 0.0346 (0.0105) & 0.0589 (0.0076) & 0.0652 (0.0243) \\
%     mirostat & 0.0325 (0.0076) & 0.0427 (0.0100) & 0.0468 (0.0058) \\
%     eta & 0.0264 (0.0201) & 0.0691 (0.0104) & 0.0610 (0.0132) \\
%     typical & 0.0691 (0.0125) & 0.0366 (0.0086) & 0.0488 (0.0173) \\
%     \end{tabular}
%     \caption{Llama-2-7B human\_eval}
%     \label{tab:my_label}
% \end{table}

\section{Conclusion}
In this work, we propose an evaluation protocol to assess the trade-off between diversity and quality
% resolved
% \mario{you use "intrinsic capacity" in the intro and conclusion ... but it is never explained what it actually means. i believe it is also neither used in the method nor the experiments.} 
of truncation sampling methods for open-ended text generation. Our evaluation enjoys the merit of being independent of parameter tuning for the curated tasks. 
The evaluation results also serve as a user reference for different downstream tasks.

\clearpage

% \textcolor{red}{move this to the discussion about the sweet spots of different sampling methods}
% For example, Top-k and Top-p have constant upper bounds, i.e., the vocabulary size $|\mathcal{V}|$ and 1, respectively. In contrast, the upper bounds of $\eta$-sampling and adaptive sampling are dependent on LLM's predicted distribution, because they truncate the tail distribution based on the likelihood of tokens and the slope of Min-Max scaled entropy, respectively. The importance of identifying the effective ranges of such parameters is also reflected in the authors' choice of numeral digit for their proposed parameters. For example, $\Delta\textit{Conf}$ is set to $0.0005$ in \citet{zhu2024improving} and $\epsilon$ is chosen from $0.0001$, $0.0009$ and etc in \citet{hewitt2022truncation}. In comparison, the $p$ values is often two digits after zero, such as $0.95$. As discussed in \Cref{sec:intro}, the parameter choice has a significant impact on the evaluation. After determining the upper and lower bounds, we normalize all the parameters to $[0, 1]$.

\clearpage

\section{Limitations}

In this work, we focus on the truncation sampling methods specially designed for the open-ended text generation scenario. There exist many related decoding strategies, which aim at improving different aspects of LLMs. For example, a line of decoding strategies is proposed to alleviate hallucination or improve the reasoning ability, e.g., Dola \cite{chuang2023dola}, Context-aware decoding \cite{shi2023trusting}, Contrastive decoding \cite{o2023contrastive} and etc. However, these methods are beyond the scope of sampling-based decoding in this study and thus not included in the discussion. Although our study is only based on text data in English for clarity, the dataset can be extended to include other languages in the future. Due to time and resource constraints, we did not include all existing sampling-based decoding methods, such as Locally Typical Sampling \cite{meister2023locally} and Min-P Sampling \cite{nguyen2024turning}, in our comparison. However, our benchmark is publicly available, and we plan to continuously update it with evaluations of additional methods in the future.

\section{Broader Impact}
Our study on the capacity of sampling methods and their appropriate parameters for open-ended text generation may further promote the application of LLMs in creative industries. There exists a potential risk that our provided findings might be abused for generating harmful or fake information. However, our study itself is neutral and the mentioned risk is a general issue that LLMs face. We call for the attention on AI-Safety in the community.

\bibliography{custom}

\clearpage
\appendix
\onecolumn

\section{Appendix}
\label{sec:appendix}

\subsection{Complete Record of the Experiment Runs}
\label{appendix:b}

\begin{table*}[ht]
    \centering
    \scriptsize
    \begin{tabular}{@{}l|ccc|ccc|ccc|ccc@{}}
      \toprule
    \multirow{3}*{Methods} & \multicolumn{9}{c|}{Evaluation Runs} &     \multicolumn{3}{c}{Mean/Std}    \\
    \cline{2-13}
     &  \multicolumn{3}{c|}{Run 1 at average Risks} &   \multicolumn{3}{c|}{Run 2 at  average Risks}   & \multicolumn{3}{c|}{Run 3 at  average Risks}  & \multicolumn{3}{c}{average Risks}
     \\
     \cline{2-13}
     & 1 & 5 & 15 &  1 & 5 & 15&  1 & 5 & 15&   1 & 5 & 15 \\
     \midrule
    Greedy Decoding &  \multicolumn{12}{c}{0.338}  
    \\
    \hline
    Naive Sampling & \multicolumn{3}{c|}{0.420} & \multicolumn{3}{c|}{0.426} & \multicolumn{3}{c}{0.416} & \multicolumn{3}{c}{0.421(0.004)} \\
    \hline
    Top-k Sampling & 0.412   &  0.447&  0.410  & 0.389 & 0.432 & 0.435 & 0.402&0.428& 0.419 &0.401(0.010)&0.436(0.008)&0.421(0.010)\\
    Top-p Sampling &  0.337  &  0.370& 0.382 &  0.367&  0.393  & 0.379& 0.362& 0.370 &0.405
    & 0.355(0.013) & 0.378(0.011) & 0.389(0.012)
    \\
    Adaptive Sampling & 0.403 & 0.416& 0.433  & 0.403& 0.416 & 0.419 & 0.378& 0.440 & 0.411
    & 0.395(0.012) & 0.424(0.011) & 0.421(0.009)
    \\
    Eta Sampling & 0.395 &0.419 &0.442& 0.387 & 0.394 &0.419&0.382&0.389&0.379
    & 0.388(0.005) & 0.401(0.013) & 0.413(0.026)
    \\
    Mirostat &     0.424 & 0.417 &0.430&0.399 &0.443&0.433& 0.415 &0.414 &0.412
    & 0.413(0.010) & 0.425(0.013) & 0.425(0.009)
    \\

\bottomrule
    \end{tabular}
        \caption{Evaluation on the TruthfulQA benchmark. Since the GPT-3 API is no longer available, we use the by the authors recommended BLEURT accuracy for comparison under the open-ended generation setup.}
           \label{tab:detail2}
\end{table*}

The scores of the individual runs on TruthfulQA benchmark are recorded in \Cref{tab:detail2}, and the means and standard errors of recalls and risks at all average Risks are listed in \Cref{tab:detail1}. Note that due to a fixed amount of computation budget, we search the corresponding parameter value for each truncation sampling method till the average risk is close enough to the predefined value, thus resulting in the variations of the average risks. However, such variations are negligible given the minor differences.

\begin{figure}[ht]
    \centering

    \begin{subfigure}[t]{0.43\textwidth}
    \centering
    \includegraphics[width=\textwidth]{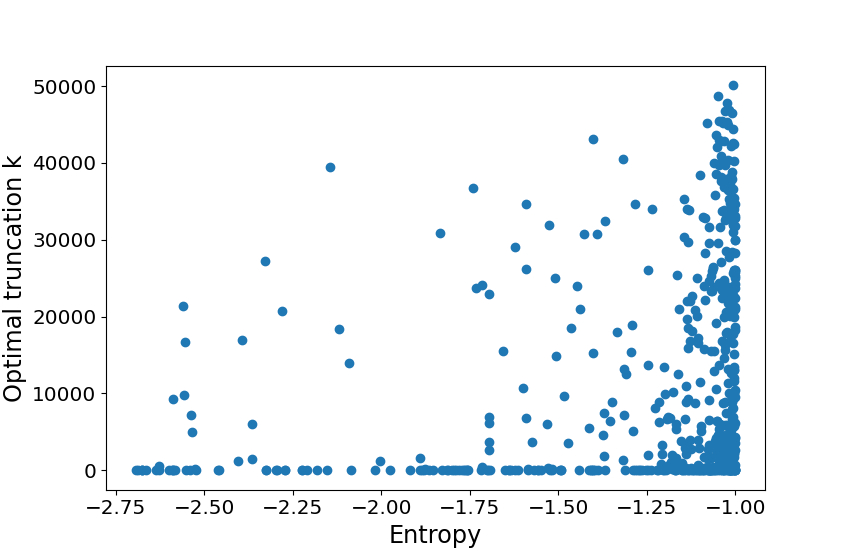}
    \caption{The Pearson's correlation is $0.24777$ for GPT2-XL.}    
    \end{subfigure}
     \begin{subfigure}[t]{0.43\textwidth}
    \centering
    \includegraphics[width=\textwidth]{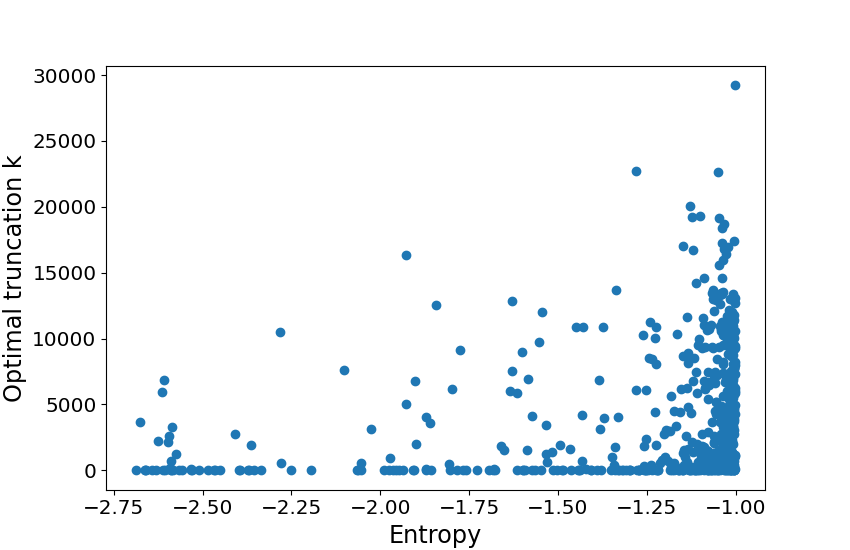}
    \caption{The Pearson's correlation is $0.24784$ for Llama-2-7B.}    
    \end{subfigure}
    \caption{Scatter plots between the entropy values and optimal truncation values.}
    \label{fig:correlation}
\end{figure}

Although Top-p sampling is indeed also adaptive regarding the truncation position, we show that Top-p sampling have a inherent limitation. When a larger portion of the probability mass is concentrated in the first few tokens (this often indicates smaller entropy), a fixed cumulative probability threshold will cut a longer tail off, and vice versa. However, there's merely a weak correlation between the entropy of the LLM's prediction and optimal truncation values, see \Cref{fig:correlation}. 

\subsection{The Advantage of Probability-Independent Metrics}
\label{appendix:a}
\begin{figure*}[ht]
    \centering
     \begin{subfigure}[t]{0.48\textwidth}
    \centering
    \includegraphics[width=\textwidth]{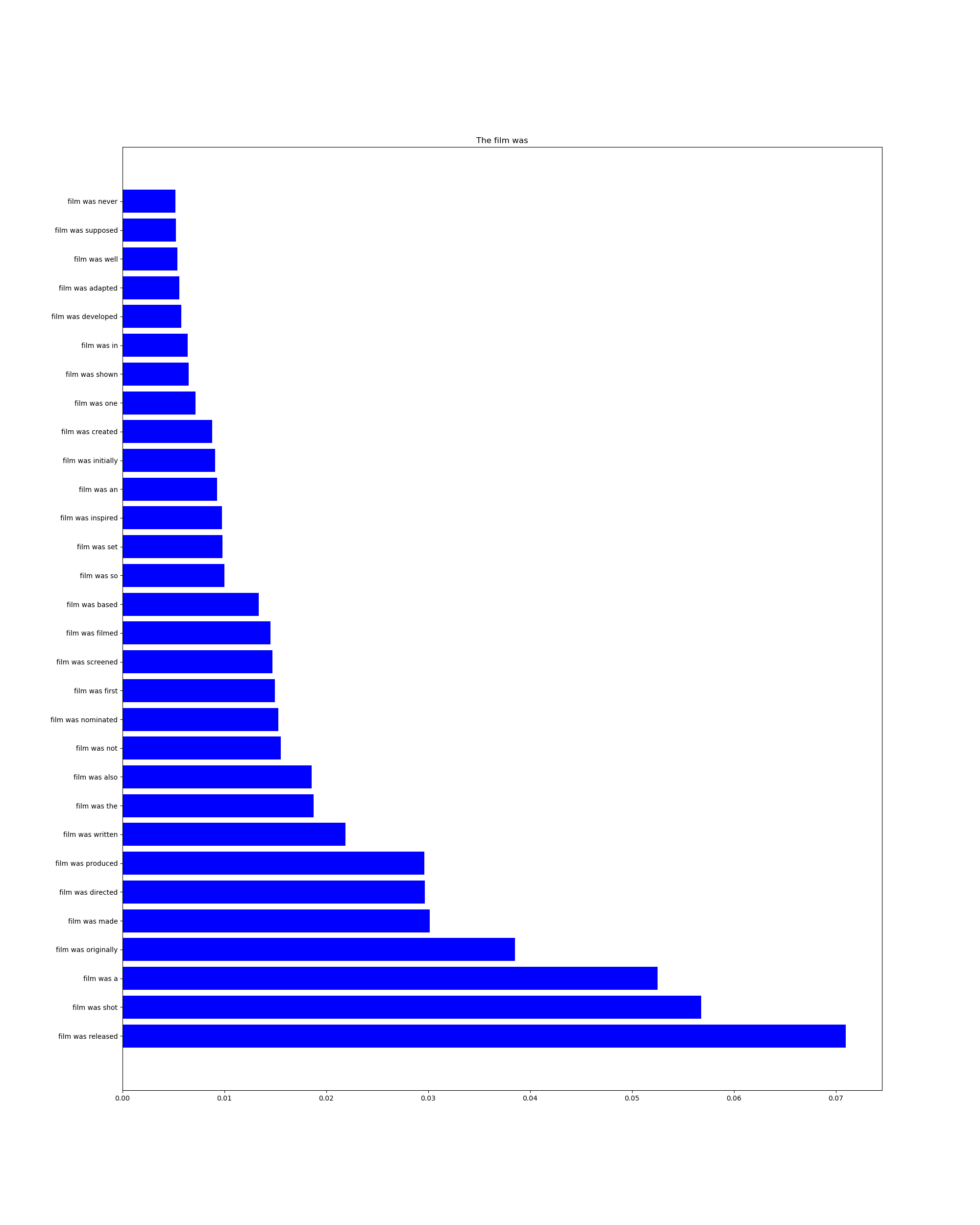}
    \caption{Top 30 by gpt2-xl predicted next candidate tokens and their corresponding likelihood given the prefix "The film was"}    
    \end{subfigure}
     \begin{subfigure}[t]{0.48\textwidth}
    \centering
    \includegraphics[width=\textwidth]{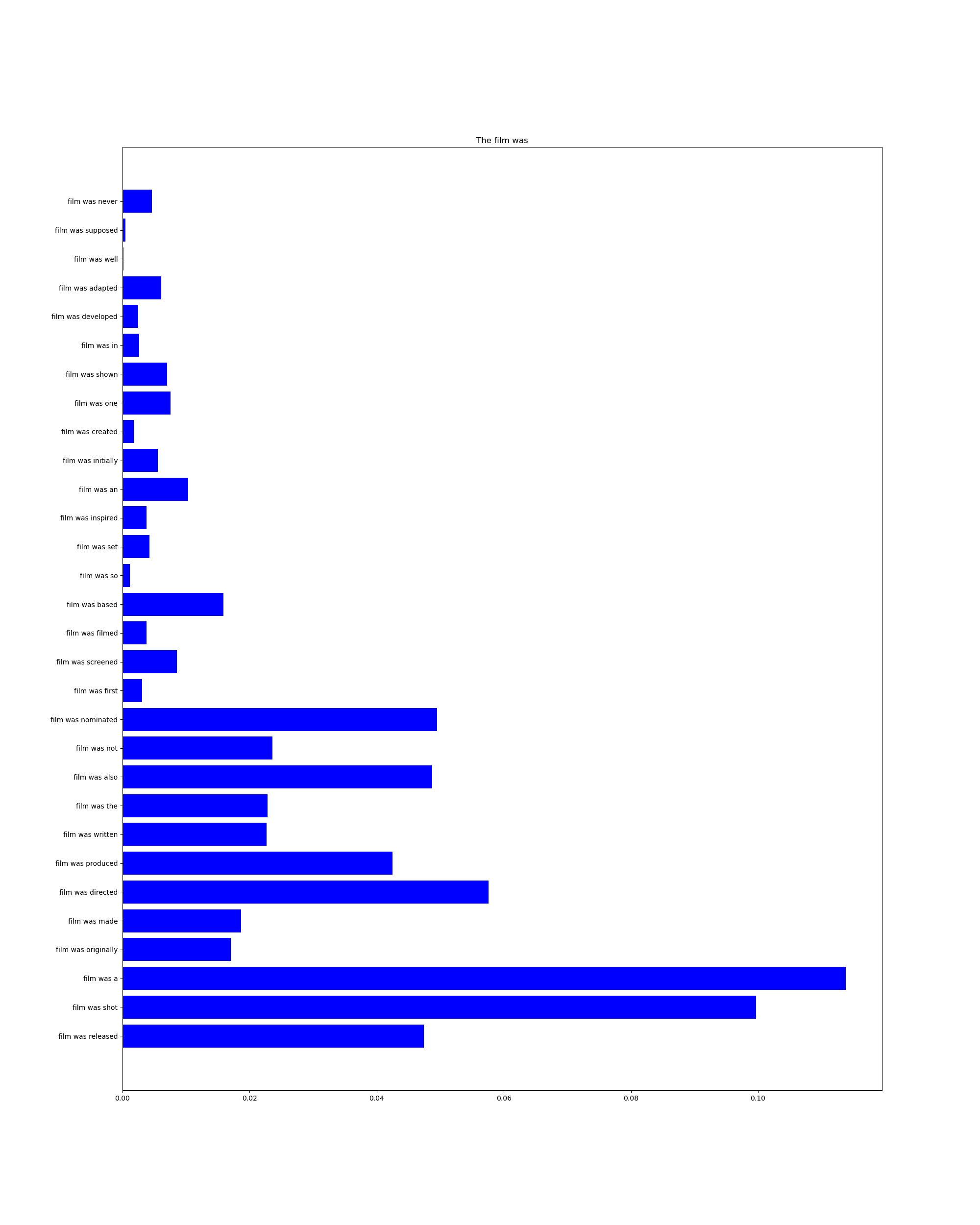}
    \caption{Top 30 by gpt2-xl predicted next candidate tokens and their corresponding empirical probability given the  prefix "The film was".}    
    \end{subfigure}
    \begin{subfigure}[t]{0.48\textwidth}
    \centering
    \includegraphics[width=\textwidth]{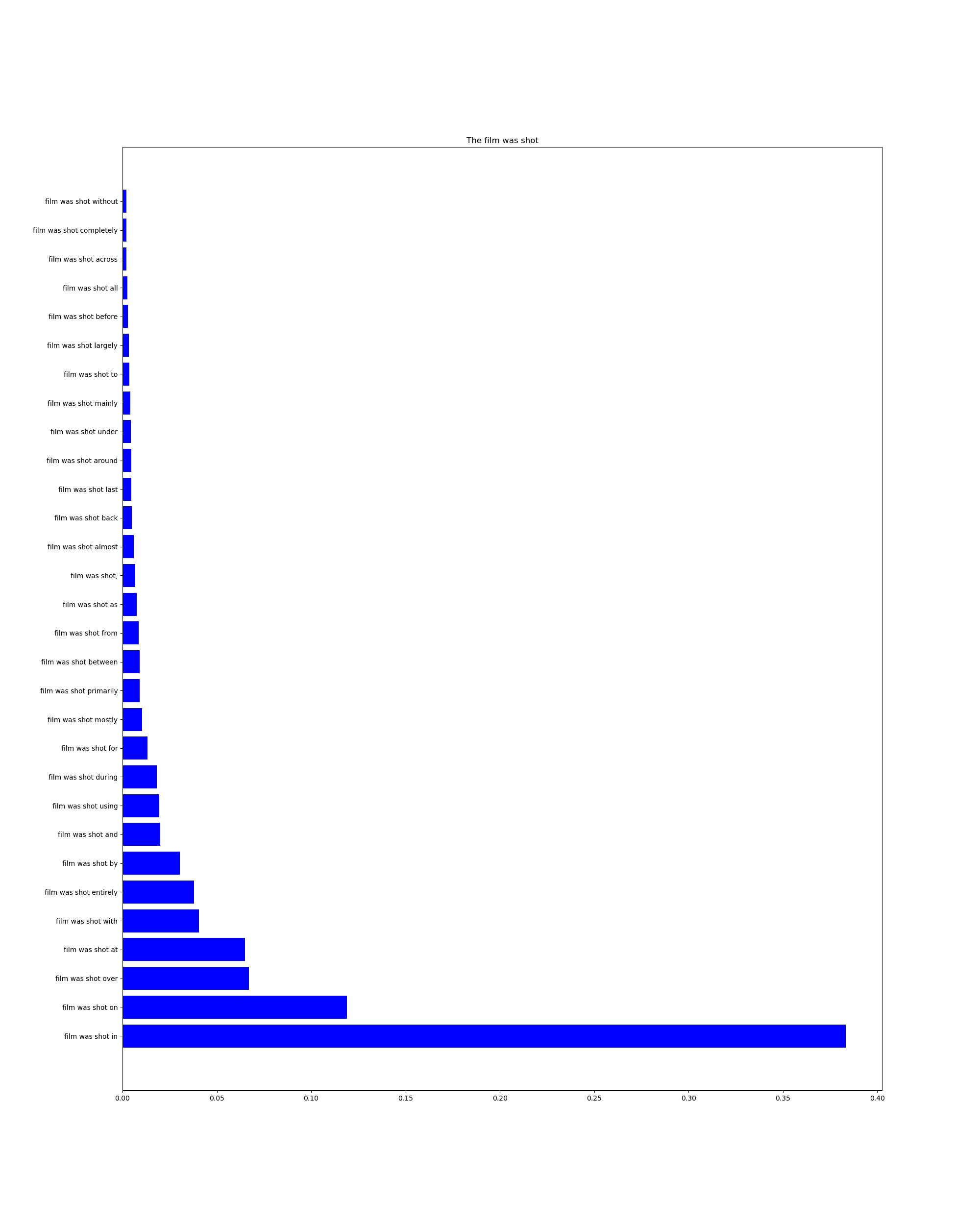}
    \caption{Top 30 by gpt2-xl predicted next candidate tokens and their corresponding likelihood given the  prefix "The film was shot".}    
    \end{subfigure}
     \begin{subfigure}[t]{0.48\textwidth}
    \centering
    \includegraphics[width=\textwidth]{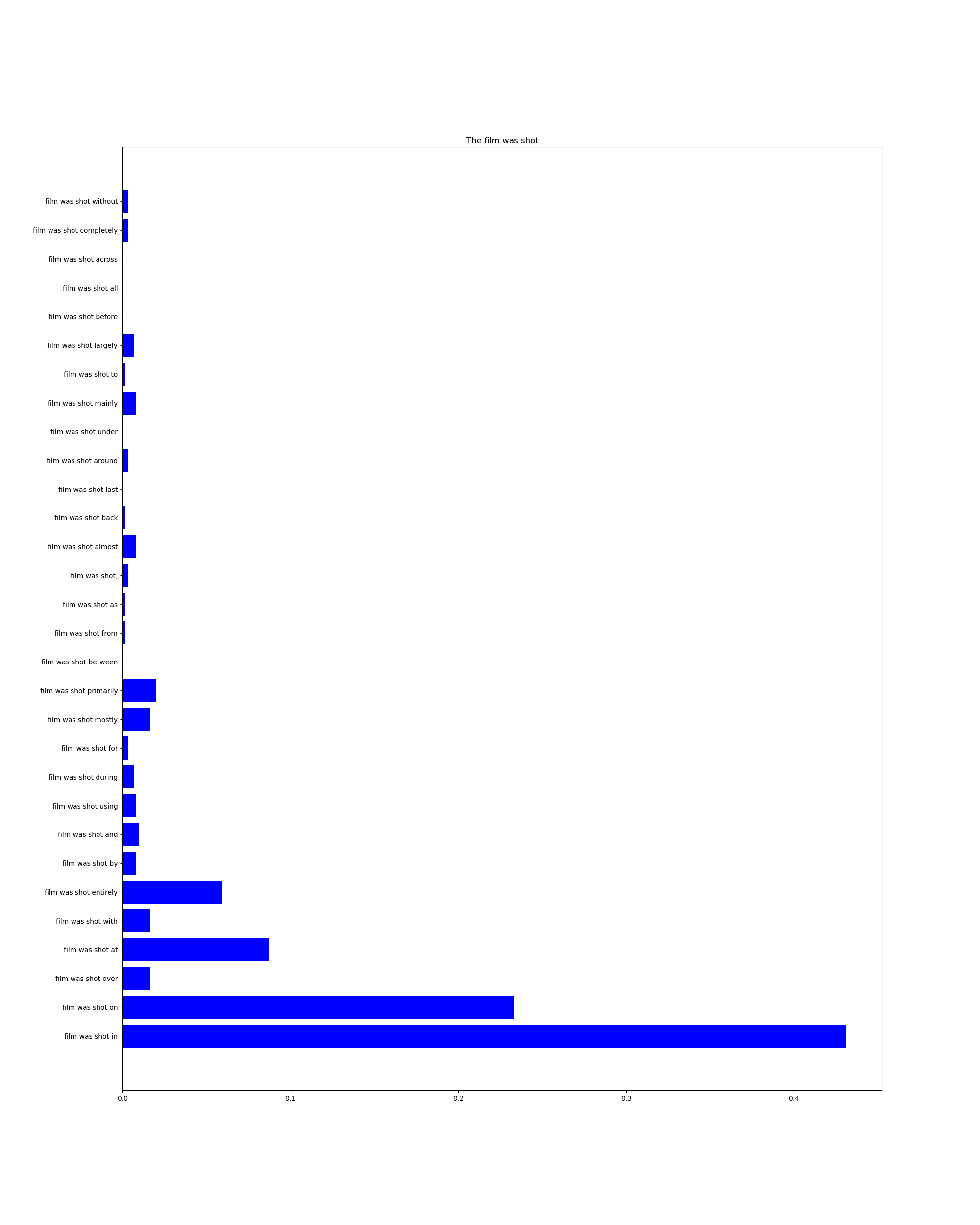}
    \caption{Top 30 by gpt2-xl predicted next candidate tokens and their corresponding empirical probability given the prefix "The film was shot".}    
    \end{subfigure}
    \caption{Comparing the probabilities predicted by gpt2-xl and calculated using the word frequencies based on our collected CP-Trie data.}
    \label{fig:llm-prediction}
\end{figure*}

In this section, we explain the practical advantages of our proposed probability-independent recall and risk metrics. As can be seen in \Cref{fig:llm-prediction}, the empirical distribution aligns with the by gpt2-xl predicted distribution given the same prefix in general: most of the tokens which posses high likelihood in the prediction also has a high probability based on the word frequencies of our collected CP-Trie data. However, there exists two differences:
\begin{itemize}
    \item Some tokens with high likelihood according to gpt2-xl have much lower probability according to the empirical distribution. The ranking of each tokens w.r.t. probability also differ in the two distributions.
    \item A few tokens which should be reasonable candidates (by manual check) have 0 probability according to the empirical distribution.
\end{itemize}

For the first issue, as discussed in \Cref{sec:issue}, there exists no ideal probabilities for each token, and the discrepancy is not solvable by simply increasing the size of the data. For example, the "perfect" probabilities of the candidate tokens "with" and "at" are undefined and could even be regarded as equivalently important for open-ended text generation.

The second difference highlights the reliability of LLMs, i.e., the tokens which are assigned high likelihoods are in most cases reasonable. Note that we ignore the risk within the estimated optimal allowed set by design: All the tokens are counted as reasonable till the last token which has non-zero empirical probability, when they are arranged in a descending order according to the predicted probabilities. Thus these tokens with zero probabilities in the empirical distribution will not affect our evaluation of risk, making our method robust to noises and insufficient data support.

\begin{table*}[ht]
    \centering
    \scriptsize
    \begin{tabular}{@{}c|ccccccccc@{}}
    \toprule
       \multirow{2}*{Method}   & \multicolumn{9}{c}{GPT2-XL} \\
          & Parameter & Risk & Recall  & Parameter & Risk & Recall & Parameter & Risk & Recall \\
              \midrule
        Top-k &   15    & 1.029 (0.006) & 0.220(0.0006) & 64  & 5.040 (0.613) & 0.290 (0.017) &    184 & 14.983(1.781) &  0.340 (0.018)
   \\
        Top-p &    0.5705 &0.999 (0.015) & 0.170 (0.0005)    &0.746& 5.011(2.129) & 0.240 (0.015) &  0.8555   & 15.022 (6.210)  & 0.338 (0.016)\\
        Adaptive  &9.5e-4&  1.000 (0.006) & 0.252 (0.0007) &0.00011& 4.997 (0.679) & 0.339(0.018)& 2.5e-05 & 14.995 (2.241) & 0.413 (0.018) \\
        Eta  &  0.318 & 1.000 (0.013)& 0.198 (0.0005) & 0.011& 4.945 (1.484) & 0.301 (0.016)&  0.001 & 14.998 (4.261) & 0.404 (0.017) \\
        Mirostat &   4.425 & 0.999 (0.005) & 0.236 (0.0007) & 5.9475& 5.001 (0.717) & 0.326 (0.018)& 6.76 & 14.982 (2.501) & 0.401 (0.018)\\
        % Typical &0.176& 1.002 (     0.009) &      0.189 (     0.014) &0.421& 5.009 (     1.503) &      0.253 (     0.015 &0.64& 15.003 (     6.867) &      0.306 (     0.016) \\
\midrule
      \multirow{2}*{Method}   & \multicolumn{9}{c}{Llama-2-7b} \\
          & Parameter & Risk & Recall  & Parameter & Risk & Recall & Parameter & Risk & Recall \\
              \midrule
    Top-k & 14 & 0.986 (0.126) & 0.226 (0.016)  & 61 & 4.987 (0.587) & 0.296 (0.017) & 177 & 14.961 (1.722) & 0.369 (0.018)
    \\ 
    Top-p & 0.54 & 0.999 (0.529) & 0.156 (0.012) & 0.7665 & 4.990 (2.331) & 0.254 (0.015) & 0.9 & 14.989 (6.208) & 0.400 (0.016)
    \\
    Adaptive &   0.0011 & 1.051 (0.154) & 0.257 (0.016) & 0.00014 & 4.991 (0.856) & 0.364 (0.017) & 3.1e-5 & 14.995 (2.966) & 0.470 (0.017)
    \\
    Eta &  0.512 & 1.000 (0.563)  & 0.192 (0.014)   & 0.023 & 5.007 (2.599) & 0.297 (0.016)& 0.002 & 13.487 (6.531) & 0.407 (0.017)
    \\
    Mirostat &  4.253 & 1.000 (0.133) & 0.236 (0.016)  & 5.82 & 4.993 (0.650) & 0.349 (0.018) & 6.628 & 15.022 (2.286) & 0.474 (0.017)\\
    % Typical & 0.1745 &   0.999 (     0.491) &      0.181 (     0.013) & 0.4705 & 5.000 (     2.255) &      0.267 (     0.015) & 0.812 &   14.998 (     6.170) &      0.360 (     0.016) \\

    \midrule
\multirow{2}*{Method}   & \multicolumn{9}{c}{Llama-3-8B} \\
          & Parameter & Risk & Recall  & Parameter & Risk & Recall & Parameter & Risk & Recall \\
              \midrule

    Top-k &  14 & 1.023 (0.128) & 0.228 (0.016) & 59 & 4.982 (0.576) & 0.290 (0.017) & 172 &     15.025 (     1.701) &      0.346 (     0.018)\\
    Top-p &   0.5395 &  1.000 (0.451) & 0.154 (0.013) &0.736 &  4.998 (2.061) & 0.224 (0.014) & 0.855 &     14.993 (     5.770) &      0.326 (     0.016)\\
    Adaptive & 0.0011 &  1.133 (0.167) & 0.260 (0.017)& 0.00017& 5.006 (0.787) & 0.343 (0.018) & 3.7e-5 &     15.007 (     2.685) &      0.418 (     0.018)\\
    Eta &  0.673 & 1.000 (0.445) & 0.181 (0.014) & 0.029 & 5.009 (2.112) & 0.271 (0.016) & 0.002 &     15.012 (     6.009) &      0.373 (     0.017)\\
    Mirostat &  4.24 & 1.001 (0.139) & 0.230 (0.016) & 5.8175 & 5.001 (0.804) & 0.318 (0.018) & 6.6925 &     14.996 (     2.630) &      0.393 (     0.018)\\
%     Typical & 0.143& 1.003 (     0.404) &      0.174 (     0.013) 
% & 0.4225& 4.996 (     1.922) &      0.247 (     0.015) & 0.6365& 15.026 (     6.239) &      0.294 (     0.016) \\
    \hline
    \multirow{2}*{Method}   & \multicolumn{9}{c}{Llama-3-70B} \\
          & Parameter & Risk & Recall  & Parameter & Risk & Recall & Parameter & Risk & Recall \\
          \hline
    Top-k & 14 & 1.014 (     0.127) &      0.230 (     0.016)& 60& 5.038 (     0.581) &      0.295 (     0.017)   & 173 &     15.024 (     1.695) &      0.352 (     0.018)\\
    Top-p &  0.5695 & 1.001 (     0.502) &      0.158 (     0.013)&   0.758  &      4.999 (     2.386) &      0.237 (     0.015)&0.8705&     14.960 (     6.685) &      0.332 (     0.016)\\
    Adaptive &  0.0011 &   1.004 (     0.137) &      0.263 (     0.017) &  0.00014 &      5.013 (     0.787) &      0.353 (     0.018)&3.16e-5&     14.986 (     2.778) &      0.424 (     0.018)
 \\
    Eta &   0.37 &   1.004 (     0.137) &      0.263 (     0.017
) & 0.014&      5.032 (     2.231) &      0.295 (     0.016)&0.001  &     15.076 (     6.265) &      0.398 (     0.018) \\
    Mirostat &  4.21 & 1.001 (     0.138) &      0.230 (     0.016
) &5.91 &      5.001 (     0.708) &      0.332 (     0.018&6.84 &     15.021 (     2.193) &      0.417 (     0.018)\\
% Typical & 0.178&   1.002 (     0.240) &      0.183 (     0.014) &0.4575& 4.998 (     2.273) &      0.252 (     0.015) 
% & 0.6605& 15.003 (     5.693) &      0.295 (     0.016) \\

\hline
 \multirow{2}*{Method}   & \multicolumn{9}{c}{Llama-2-70b} \\
          & Parameter & Risk & Recall  & Parameter & Risk & Recall & Parameter & Risk & Recall \\
          \hline
        Top-k & 14 &  1.002 (     0.128) &      0.232 (     0.016
) & 60 &      4.982 (     0.583) &      0.307 (     0.017)
& 174 &     14.964 (     1.712) &      0.375 (     0.018)
\\
        Top-p & 0.6535 &  0.999 (     0.475) &      0.189 (     0.013
) & 0.8465 &      4.988 (     2.136) &      0.316 (     0.016) 
& 0.9395 &     15.019 (     5.522) &      0.468 (     0.016)
\\
        Adaptive & 0.0011 &  1.000 (     0.142) &      0.269 (     0.017
 ) & 1.2e-4&      4.995 (     0.796) &      0.374 (     0.017)
 & 2.3e-5 &     15.007 (     2.697) &      0.485 (     0.017)
 \\
        Eta & 0.092 &   1.002 (     0.304) &      0.236 (     0.015
) & 0.003 &      5.057 (     1.590) &      0.378 (     0.017)
& 0.00021 &     15.001 (     4.243) &      0.510 (     0.017)
\\
        Mirostat & 4.16 & 1.001 (     0.135) &      0.238 (     0.016
 & 5.7875 &      5.004 (     0.684) &      0.353 (     0.018)
 &6.67 &     14.991 (     2.125) &      0.478 (     0.017)
 \\
 % Typical &0.3&   1.001 (     0.305) &      0.217 (     0.014) & 0.603&  5.005 (     2.120) &      0.291 (     0.016) &0.9345&  14.963 (     5.314) &      0.467 (     0.016) \\
        \hline
    \multirow{2}*{Method}   & \multicolumn{9}{c}{Mixtral-8x7B} \\
          & Parameter & Risk & Recall  & Parameter & Risk & Recall & Parameter & Risk & Recall \\
          \hline
        Top-k & 15 &  1.028 (     0.134) &      0.229 (     0.016) & 63 &      4.978 (     0.598) &      0.301 (     0.017) 
        & 183   &     14.967 (     1.757) &      0.366 (     0.018)
        \\
        Top-p & 0.6505 &   1.000 (     0.535) &      0.192 (     0.014 ) & 0.8375 &      5.007 (     2.423) &      0.303 (     0.015) 
        & 0.9325 &     14.966 (     6.139) &      0.456 (     0.016)
        \\
        Adaptive &  0.00105 &  1.000 (     0.148) &      0.265 (     0.017
 ) & 0.00011 &      4.994 (     0.798) &      0.372 (     0.018)
 & 2.1e-5 &     15.014 (     2.802) &      0.476 (     0.017)
 \\
        Eta & 0.087 &    1.001 (     0.335) &      0.241 (     0.015
) & 0.003  &      5.061 (     1.822) &      0.375 (     0.017)
& 0.000215 &     14.991 (     4.922) &      0.506 (     0.017)
\\
        Mirostat & 4.2775 &  1.000 (     0.143) &      0.238 (     0.016) & 5.845 &      4.995 (     0.710) &      0.346 (     0.018)
        & 6.6875 &     14.998 (     2.213) &      0.461 (     0.018)
        \\
        % Typical &0.302&  0.999 (     0.344) &      0.217 (     0.014) &0.5965& 4.995 (     2.111) &      0.289 (     0.015) &0.9225& 14.992 (     5.351) &      0.451 (     0.017) \\
        \hline
   
    \multirow{2}*{Method}   & \multicolumn{9}{c}{Mistral-7B} \\
          & Parameter & Risk & Recall  & Parameter & Risk & Recall & Parameter & Risk & Recall \\
          \hline
        Top-k &  14  &      0.965 (     0.126) &      0.224 (     0.016) & 62 &      4.968 (     0.596) &      0.297 (     0.017)
        & 181 &     15.006 (     1.759) &      0.364 (     0.018)
        \\
        Top-p & 0.6565 &  1.001 (     0.539) &      0.194 (     0.014) & 0.8375 &      4.996 (     2.476) &      0.303 (     0.016
) 
& 0.9315 &     15.038 (     6.315) &      0.447 (     0.016)
\\
        Adaptive & 0.00105  &      1.001 (     0.152) &      0.260 (     0.016) & 0.000115 &      4.993 (     0.809) &      0.364 (     0.018)
        & 2.2e-5  &     14.999 (     2.757) &      0.466 (     0.017)
        \\
        Eta & 0.075 &      0.997 (     0.307) &      0.243 (     0.015) & 0.003 &      4.640 (     1.542) &      0.368 (     0.017) 
        & 0.000196 &     15.009 (     4.712) &      0.505 (     0.017)
        \\
        Mirostat &4.1825 &      1.000 (     0.141) &      0.236 (     0.016) & 5.8125 &      4.999 (     0.721) &      0.345 (     0.018) 
        & 6.71 &    14.978 (     2.213) &      0.468 (     0.018)
        \\
        % Typical &0.302&  1.000 (     0.251) &      0.215 (     0.014) &0.577&      4.998 (     1.278) &      0.283 (     0.015) &0.927&   14.990 (     5.047) &      0.453 (     0.016)\\
        
        \bottomrule
    \end{tabular}
    \caption{Critical Parameters of different truncation sampling methods at different average Risks using different models.}
    \label{tab:detail1}
\end{table*}

\end{document}